\documentclass[letterpaper]{article} 
\usepackage{aaai2026}  
\usepackage{times}  
\usepackage{helvet}  
\usepackage{courier}  
\usepackage[hyphens]{url}  
\usepackage{graphicx} 
\usepackage{natbib}  
\usepackage{caption} 
\usepackage{amsmath}
\usepackage{amsfonts}
\usepackage{booktabs}
\usepackage{multirow} 
\usepackage{pgfplots}
\pgfplotsset{compat=1.18}
\usepackage{algorithm}
\usepackage{algorithmic}

\usepackage{newfloat}
\usepackage{listings}
\DeclareCaptionStyle{ruled}{labelfont=normalfont,labelsep=colon,strut=off} 
\floatstyle{ruled}
\newfloat{listing}{tb}{lst}{}
\floatname{listing}{Listing}
\title{CtrlFuse: Mask-Prompt Guided Controllable Infrared and Visible Image Fusion}
\author {
    Yiming Sun\textsuperscript{\rm 1},
    Yuan Ruan\textsuperscript{\rm 1},
    Qinghua Hu\textsuperscript{\rm 2},
    Pengfei Zhu\textsuperscript{\rm 1,3,4}\thanks{Corresponding author}
}
\affiliations {
    \textsuperscript{\rm 1}School of Automation, Southeast University, Nanjing, China\\
    \textsuperscript{\rm 2}School of Artificial Intelligence, Tianjin University, Tianjin, China\\
    \textsuperscript{\rm 3}Low-Altitude Intelligence Laboratory,
Xiong'an National Innovation Center, Xiongan, China\\
    \textsuperscript{\rm 4}Xiong'an Guochuang Lantian Technology Co., Ltd., Xiongan, China\\
    \{sunyiming, ruanyuan\}@seu.edu.cn, \{huqinghua, zhupengfei\}@tju.edu.cn
}

\usepackage{bibentry}

\begin{document}

\maketitle

\begin{abstract}
Infrared and visible image fusion generates all-weather perception-capable images by combining complementary modalities, enhancing environmental awareness for intelligent unmanned systems. Existing methods either focus on pixel-level fusion while overlooking downstream task adaptability or implicitly learn rigid semantics through cascaded detection/segmentation models, unable to interactively address diverse semantic target perception needs. We propose CtrlFuse, a controllable image fusion framework that enables interactive dynamic fusion guided by mask prompts. The model integrates a multi-modal feature extractor, a reference prompt encoder (RPE), and a prompt-semantic fusion module (PSFM). The RPE dynamically encodes task-specific semantic prompts by fine-tuning pre-trained segmentation models with input mask guidance, while the PSFM explicitly injects these semantics into fusion features. Through synergistic optimization of parallel segmentation and fusion branches, our method achieves mutual enhancement between task performance and fusion quality. Experiments demonstrate state-of-the-art results in both fusion controllability and segmentation accuracy, with the adapted task branch even outperforming the original segmentation model. 
\end{abstract}

\begin{links}
\link{Code}{https://github.com/Sevryy/CtrlFuse}
\end{links}
\section{Introduction}
Infrared and visible image fusion aims to combine complementary information from both modalities to generate comprehensive representations of scenes~\cite{ma2016infrared, Tang2024Mask-DiFuser}, thereby enhancing the environmental perception capabilities of intelligent unmanned systems from night to day~\cite{sun2024dynamic,pmlrsun25k}. Although visible images provide rich color information and high spatial resolution, their performance in downstream tasks degrades under poor illumination conditions. Infrared images effectively compensate for the limitations of visible imaging in darkness through thermal target imaging, but lack texture information of targets, leading to misidentification issues in downstream applications. How to better leverage the advantages of both imaging modalities to improve various downstream applications has become a key research focus.

\begin{figure}[htbp] 
    \centering
    \includegraphics[width=0.45\textwidth]{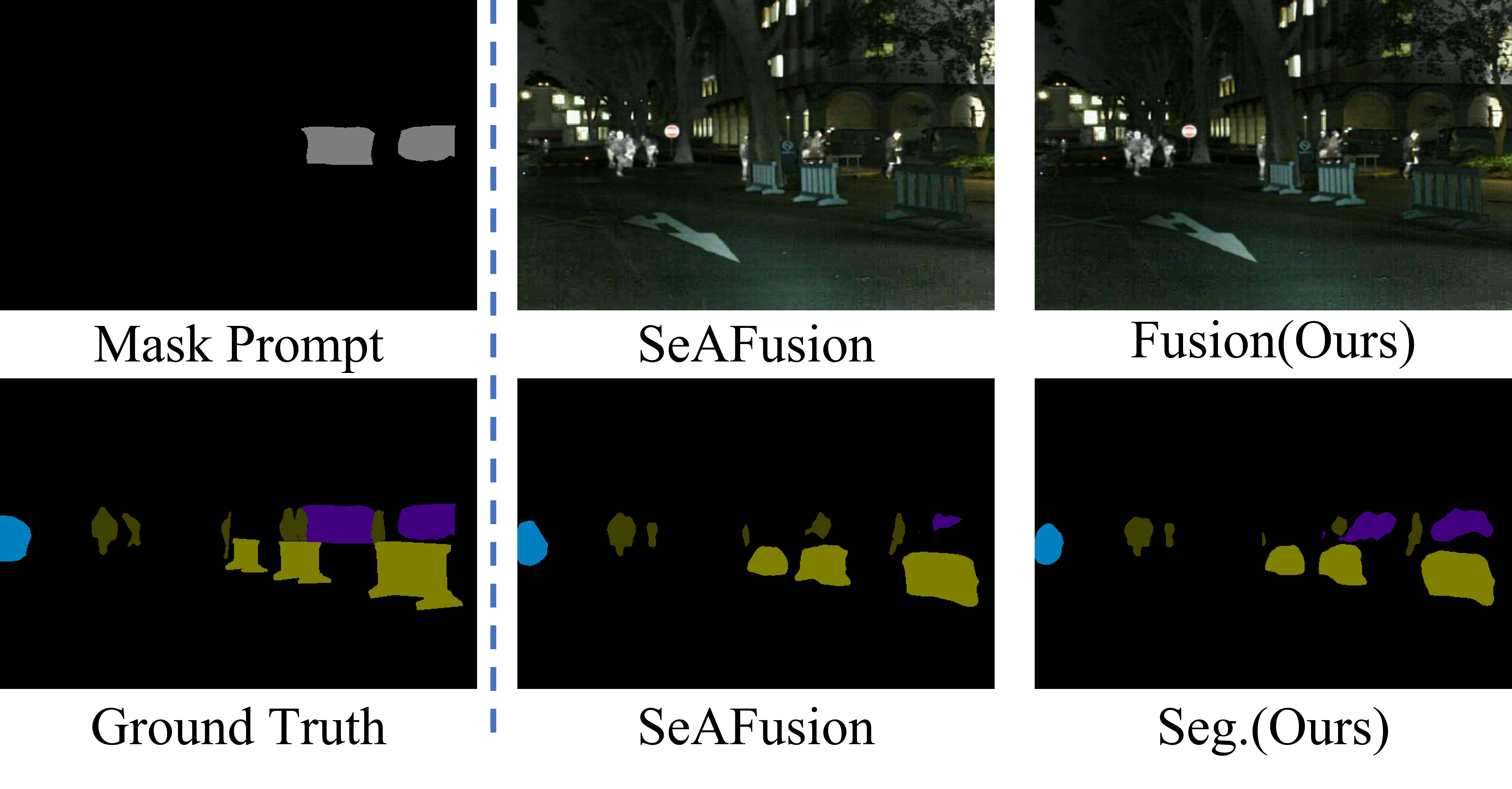}
    \caption{The importance of mask prompt-based interactive controllable image fusion for the performance of downstream application models (e.g., semantic segmentation).
    }
    \label{fig1}
\end{figure}

With the development of deep learning, several representative infrared and visible image fusion methods have emerged~\cite{cao2023multi}. Autoencoder-based methods enhance fusion performance through multi-scale feature decomposition or pretraining-finetuning strategies. Generative adversarial network (GAN)-based~\cite{ma2020ddcgan} approaches preserve effective information from different modalities by establishing adversarial learning between fused images and multimodal source images. Beyond CNN-based methods, transformer architectures and diffusion models~\cite{Tang2024DRMF} have also attracted researchers' attention. Traditional image fusion tasks primarily focus on pixel-level consistency between fused and source images, evaluating performance through image quality assessment metrics, while neglecting the crucial requirement that fused images should effectively serve downstream perception tasks for improved task performance.

In recent years, researchers have recognized this need, leading to the proposal of task-related multimodal image fusion methods~\cite{Tang2024C2RF,liu2024promptfusion}. These approaches connect image fusion networks with downstream task networks (e.g., object detection or semantic segmentation models) through joint optimization, enabling fusion models to implicitly learn semantically relevant information for downstream tasks. However, this paradigm severely restricts the fusion model's semantic perception capability to the predefined recognition types of the cascaded downstream task model, failing to dynamically control the attention to specific targets according to varying application requirements. As shown in Fig.~\ref{fig1}, although existing methods have learned target semantics during training, they still struggle to adapt to real-world vehicle segmentation scenarios. Our method significantly enhances the semantic segmentation capability for designated targets (Car) through controllable prompt guidance, improving the practicality of task-driven image fusion models. Therefore, constructing semantically controllable multimodal image fusion architectures that enable dynamic controllable fusion according to different semantic requirements could bridge high-level downstream applications with low-level image fusion tasks.

In this paper, we propose a controllable image fusion method with multimodal semantic-aware prompt tuning (CtrlFuse), establishing a semantic-guided controllable multimodal image fusion framework through prompt tuning of the foundation model with superior semantic perception capability (Segment Anything Model~\cite{kirillov2023segment}, SAM). Specifically, the proposed CtrlFuse consists of four components: a multimodal backbone encoder-decoder, a reference prompt encoder (RPE), a prompt semantic fusion module (PSFM), and the pre-trained SAM. In the proposed RPE, we construct support features and query features that form prompt features for SAM fine-tuning under mask guidance. The PSFM explicitly fuses semantic prompt features, segmentation masks from SAM, and multimodal features from the encoder, achieving dynamic explicit injection of semantic information. Our method enhances performance on multimodal image fusion tasks by jointly optimizing the loss functions for both the SAM task branch and the image fusion branch. The main contributions are summarized as follows:
\begin{itemize}
\item We propose an interactively controllable multimodal image fusion framework that establishes a concise controllable fusion paradigm through dynamic mask prompts and explicit fusion of multimodal features with foundation model fine-tuning.
\item We design a reference prompt encoder to dynamically generate semantic prompts for SAM fine-tuning, and develop a prompt semantic fusion module to explicitly aggregate semantic prompts with multimodal features, enhancing semantic perception capability.
\item Extensive experiments validate the superior fusion performance and semantic controllability of our method. In particular, our approach achieves enhanced fusion capability through multimodal semantic fine-tuning, revealing the synergistic advantage of mutual promotion between the fusion and segmentation tasks.
\end{itemize}

\begin{figure*}[!t]
    \centering
    \includegraphics[width=0.95\textwidth]{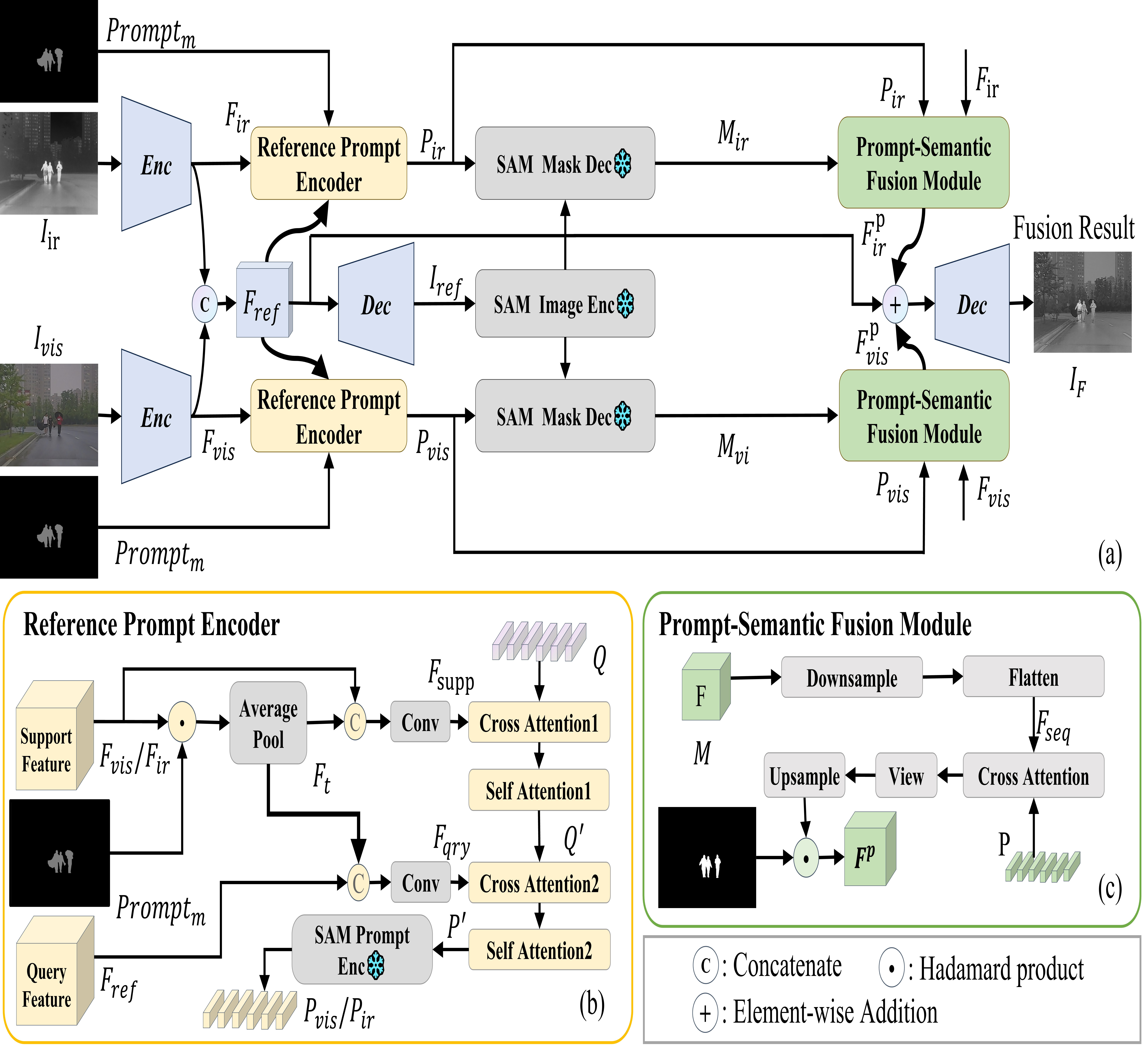} 
    \caption{The architecture of CtrlFuse. CtrlFuse consists of two reference prompt encoders, two Prompt-Semantic fusion modules guided by the prompt mask, a set of infrared and visible feature encoders, and the auxiliary network.}
    \label{fig:Framework}
\end{figure*}

\section{Related Works}
{\bf Infrared and Visible Image Fusion} integrates complementary multi-modal features to preserve salient information from both sources~\citep{ma2019infrared,zhang2021image,10812907}. Ma et al.~\cite{ma2019fusiongan} proposed FusionGAN to formulate the image fusion task as an adversarial game between preserving infrared thermal radiation and visible texture details. Zhao et al.~\cite{ZhaoDIDFuse2020} proposed a deep image decomposition framework that separates source images into background and detail feature maps, followed by dedicated fusion strategies for each feature. 
Li et al.~\cite{li2020nestfuse} proposed a fusion network featuring nested connectivity to achieve multi-scale fusion of multimodal images. 
Recently, some task-driven methods have also attracted wide attention. Sun et al.~\cite{sun2022detfusion} proposed a detection-driven network for infrared and visible image fusion that leverages target-specific features from object detection tasks to guide the fusion process. Liu et al.~\cite{liu2022target} proposed a fusion framework that enhances object detection performance through dual adversarial optimization. Tang et al.~\cite{tang2022image} proposed a semantic-aware fusion network that cascades the fusion module with a semantic segmentation module to enhance fusion performance. Yi et al.~\cite{yi2024text} proposed a text-guided image fusion framework that utilizes prompts to enable degradation-aware fusion. However, existing task-driven methods typically cascade image fusion with downstream application tasks, with high-level visual models (detection or segmentation) guiding the optimization of fusion models. This approach only supports the learning of fixed semantic categories and is difficult to adapt dynamically to complex and diverse semantic category perception requirements. Furthermore, it cannot achieve interactive and controllable fusion.

{\bf Interactive Deep Learning Model} refers to a class of neural network architectures that enable human-steered regulation of model behavior~\cite{cao2024controllable, Tang2025ControlFusion}. Kirillov et al.~\cite{kirillov2023segment} proposed the Segment Anything Model (SAM), a general-purpose segmentation system featuring interactive control and strong zero-shot generalization. Based on SAM, many other interactive large models have also been proposed, such as VRP-SAM~\cite{sun2024vrp}, GroundedSAM~\cite{ren2024grounded}, and SAGE~\cite{wu2025every}. In recent years, diffusion models have demonstrated significant progress in controllability~\citep{shi2024dragdiffusion,zhang2024transparent,hoe2024interactdiffusion}. Avrahami et al.~\cite{avrahami2022blended} developed blended diffusion by effectively integrating CLIP's~\cite{radford2021learning} semantic understanding capabilities with diffusion models to achieve text-driven image inpainting. Huang et al.~\cite{huang2023composer} proposed a novel paradigm for composable conditional image synthesis that achieves creative controllability while maintaining generation quality. However, existing fusion methods predominantly focus on optimizing quantitative metrics for static outputs, while largely neglecting the development of controllable dynamics. This oversight results in limited adaptability to real-world scenarios requiring interactive adjustment or dynamic responses. Considering that image fusion should ultimately enhance downstream task performance, controllable region-aware fusion enhancement methods would provide greater practical value.

\section{Methods}
\subsection{Overall Architecture}
In this paper, we propose a semantic-aware framework enabling controllable fusion via multi-modal prompt tuning, termed CtrlFuse. In Fig~\ref{fig:Framework}, CtrlFuse contains a Reference Prompt Encoder, a Prompt-Semantic Fusion Module, and the auxiliary network.

In Fig.~\ref{fig:Framework}(a), we send a pair of infrared image $I_{ir} \in \mathbb{R}^{1 \times H \times W}$ and visible image $I_{vis} \in \mathbb{R}^{3 \times H \times W}$ into the infrared and visible encoders to extract the features $F_{ir}$ and $F_{vis}$, respectively. The structure of encoders follows~\cite{tang2022image}. Then we concatenate $F_{ir}$ and $F_{vis}$ into $F_{ref}$ as the input of the image decoder consisting of convolutional layers and activation layers to get the reference image $I_{ref}$. We send a mask as a prompt into Reference Prompt Encoder along with $F_{ir}$ or $F_{vis}$ and $F_{ref}$. In the Reference Prompt Encoder, we use $F_{i r}$ or $F_{vis}$ as support feature, $F_{ref}$ as query feature and get prompt feature embedding $P_{i r}$ and $P_{vis}$, respectively. We then send $I_{ref}$ into the frozen SAM image encoder, and the result is sent to frozen SAM mask decoder with $P_{i r}$ and $P_{vis}$ respectively, after that we get two prediction masks $M_{ir}$ and $M_{vis}$. To obtain the dual-light prompt features $F_{ir}^{\mathrm{p}}$ and $F_{vis}^{\mathrm{p}}$, we input the prediction masks, prompt embeddings, and encoded features into the Prompt-Semantic Fusion Module. Through element-wise addition of the preliminary fusion feature $F_{ref}$ with the prompt features $F_{ir}^{\mathrm{p}}$ and $F_{vis}^{\mathrm{p}}$, the final fusion features are obtained. These final features are subsequently input into the image decoder to generate the ultimate fused image $I_{\mathcal{F}} \in \mathbb{R}^{1 \times H \times W}$. 

We train the proposed CtrlFuse model in an end-to-end manner. The model is optimized mainly by calculating both the fusion loss $\mathcal{L}_{\text {fusion }}$ and the segmentation loss $\mathcal{L}_{\text {seg}}$. 

\subsection{Reference Prompt Encoder}
As shown in Fig.~\ref{fig:Framework}(b), we propose a Reference Prompt Encoder to dynamically extract prompt embeddings from the controllable regions of interest.

Taking the infrared reference prompt encoder as an example, we take $F_{ir}$ as a support feature, $F_{ref}$ as a query feature, and input them along with $Prompt_{m}$ into the encoder. To enhance the features in the regions of interest, we compute the Hadamard product between $Prompt_{m}$ and $F_{ir}$, followed by an average pooling operation to get $F_{t}$. This process can be described as follows:
\begin{equation}
F_{\text{t}} = \operatorname{AveragePool} \left( \text{Prompt}_{\text{m}} \cdot F_{\text{ir}} \right).
\end{equation}

In order to achieve a holistic feature representation that captures both detailed local information and global context, which enhances the model's performance and generalization ability, we separately concatenate $F_{ir}$ and $F_{ref}$ with $F_{t}$, and then apply convolutional layers to produce the resulting features $ F_{\text{supp}} \in \mathbb{R}^{C \times H \times W} $ and$ F_{\text{qry}} \in \mathbb{R}^{C \times H \times W} $. This process can be described as follows:
\begin{equation}
F_{supp} = \operatorname{Conv}(\operatorname{Concat}(F_{\text {ir}}, F_{\text {t}})),
\end{equation}
\begin{equation}
F_{qry} = \operatorname{Conv}(\operatorname{Concat}(F_{\text {ref}}, F_{\text {t}})).
\end{equation}

We get a set of learnable queries $Q \in \mathbf{R}^{N \times C}$, that extract different aspects of information from $F_{supp}$, where $N$ corresponding to the number of prominent features the model focuses on, is set to 40 as determined by hyperparameter experiments. We first pass the queries $Q$ and $F_{supp}$ through a cross-attention layer, and then feed the resulting output into a self-attention layer to obtain queries that are controllable for specific categories in the support image $Q^{\mathrm{'}}$. Subsequently, $Q^{\mathrm{'}}$ and $F_{qry}$ are processed through a cross-attention mechanism. The resulting output is then fed into a self-attention layer to derive a set of reference prompts $P^{\mathrm{'}}$, which correspond to specific categories present in the query image. This process can be described as follows:
\begin{equation}
Q^{\prime}=\operatorname{SelfAttn}_1\left(\operatorname{CrossAttn}_1\left(Q, F_{supp}\right)\right)
\end{equation}
\begin{equation}
P^{\prime}=\operatorname{SelfAttn}_2\left(\operatorname{CrossAttn}_2\left(Q^{\prime}, F_{qry}\right)\right)
\end{equation}

Then, $P^{\mathrm{'}}$ is passed through the frozen SAM Prompt Encoder to generate the final prompt feature embedding $P$.

\subsection{Prompt-Semantic Fusion Module}
As shown in Fig.~\ref{fig:Framework}(c), we propose a Prompt-Semantic Fusion Module to obtain category-specific prompt features $F_{ir}^{\mathrm{p}}$ and $F_{vis}^{\mathrm{p}}$. In both the infrared and visible branches, the initial encoded features $F$, the prompt feature embeddings $P$, and the corresponding SAM segmentation outputs $M$ are used as input representations. Given the notable differences in how various modalities contribute to semantic information, we employ segmentation results derived from each modality-specific branch (e.g., infrared and visible) as masks. Specifically, the quality of the segmentation outcomes directly reflects the richness and quality of the high-level features provided by that modality, thereby indicating its potential contribution to overall task performance. Such a strategy not only facilitates the quantification of each modality's effectiveness but also provides more precise guidance for multi-modal data fusion. First, we downsample the feature map $F\in \mathbb{R}^{H \times W \times C}$ to reduce the computational load of the network. Then, we flatten the downsampled features into a sequence format $F_{seq}\in \mathbf{R}^{(H*W) \times C}$. This process can be described as follows:
\begin{equation}
F_{seq}=Flatten(Down(F))
\end{equation}

We process the serialized features $F_{seq}$ together with the previously derived prompt embeddings $P$ through a cross-attention mechanism. Subsequently, the attended features are subjected to a view transformation operation to revert to their original spatial dimensions, followed by an upsampling step to restore them to the original resolution, thereby obtaining enhanced spatial features. Finally, these enhanced spatial features are element-wise multiplied with the corresponding segmentation masks to derive the enhanced features specific to the indicated categories . This process can be described as follows:
\begin{equation}
F^p=M\cdot(Up(View(CrossAttn(F_{seq},P))))
\end{equation}


\section{Experiments}
\subsection{Experimental Setting}
\noindent \textbf{Implementation Details.}
We performed experiments on a computing platform with four NVIDIA GeForce RTX 3090 GPUs. We used Adam Optimization to update the overall network parameters with the learning rate of $1.0\times 10^{-4}$. The training epoch is set to $150$ and the batch size is $4$.

\noindent \textbf{Datasets and Partition Protocol.} We conducted experiments on three publicly available datasets: FMB~\cite{liu2023multi}, MSRS~\cite{tang2022piafusion}, and DroneVehicle~\cite{Sun2022DroneBasedRC}. FMB and MSRS contain labels for semantic segmentation, and we generate masks for different categories from these labels. FMB contains $1,500$ infrared-visible image pairs captured by onboard cameras. We used $1,020$ pairs of images for training, $280$ pairs of images for validation, and the remaining $200$ pairs for evaluation. MSRS contains $1,444$ infrared-visible image pairs captured by onboard cameras. We used $1,072$ image pairs for training, $150$ image pairs for validation, and $200$ image pairs for evaluation. For DroneVehicle, we generate masks using other semantic segmentation models. We evaluated $200$ image pairs from DroneVehicle. 
We present both qualitative and quantitative analyses of the MSRS dataset in the supplementary material.




\noindent \textbf{Competing Methods.}
We compared $8$ state-of-the-art methods on three publicly available datasets. In these comparison methods, LDfusion~\cite{wang2024infrared} is the CLIP-based image fusion method,
NestFuse~\cite{li2020nestfuse} is the autoencoder-based method, DIDFuse~\cite{ZhaoDIDFuse2020} and CDDFuse~\cite{zhao2023cddfuse} are the deep learning-based image decomposition methods. SwinFuse~\cite{SwinFuse2022} is a Transformer-based method.
SeAFusion~\cite{liu2022target} and SDCFusion~\cite{liu2024semantic} are the segmentation-driven methods. PSFusion~\cite{TANG2023PSFusion} is a fusion method driven by high-level vision tasks.

\noindent \textbf{Evaluation Metrics.}
We evaluated the performance of the proposed method based on qualitative and quantitative results. The qualitative evaluation is mainly based on the visual effect of the fused image. A good fused image needs to have complementary information from multi-modal images.
The quantitative evaluation mainly uses quality evaluation metrics to measure the performance of image fusion. We selected $6$ popular metrics, including the MSE, PSNR, $N_{abf}$ , gradient-based similarity measurement ($Q_{abf}$)~\cite{Xydeas2000ObjectiveIF},SSIM~\cite{wang2004image} and SCD. We also evaluate the performance of the different methods on the typical downstream task, infrared-visible object detection, and semantic segmentation.

\begin{table}[t] 
    \centering
    \setlength{\tabcolsep}{2pt}  
    \begin{tabular}{lcccccc}
        \toprule
        Methods & MSE & PSNR & $Q_{\text{abf}}$ & $N_{\text{abf}}$ & SSIM & SCD \\
        \midrule
        \multicolumn{7}{c}{FMB Dataset} \\
        \midrule
        LDFusion             & 0.061 & 60.71 & 0.51 & 0.112 & 0.514 & 1.549 \\
        SwinFuse       & \textbf{\underline{0.042}} & 62.334 & 0.577 & \underline{0.029} & 0.905 & 1.9 \\
        NestFuse       & \underline{0.046} & 61.96 & 0.483 & 0.042 & 0.787 & 1.594 \\
        CDDFuse            & 0.048 & \textbf{62.696} & \underline{0.674} & \textbf{0.026} & \textbf{\underline{1.002}} & \textbf{1.626} \\
        DIDFuse            & 0.047 & 61.565 & 0.528 & 0.042 & 0.765 & \textbf{1.824} \\
        SeAFusion      & 0.047 & \underline{62.539} & 0.654 & \underline{0.029} & \textbf{0.964} & 1.62 \\
        PSFusion       & 0.051 & 61.517 & 0.627 & 0.056 & 0.836 & \textbf{\underline{1.875}}\\
        SDCFusion      & 0.048 & 62.456 & \textbf{0.693} & 0.031 & 0.906 & \underline{1.657}\\
        CtrlFuse(Ours)   & \textbf{0.043} & \textbf{\underline{63.292}} & \textbf{\underline{0.719}} & \textbf{\underline{0.024}} & \underline{0.925} & 1.522 \\
        \midrule
        \multicolumn{7}{c}{DroneVehicle Dataset} \\
        \midrule
        LDFusion             & 0.076 & 59.573 & 0.376 & 0.054 & 0.568 & 1.38 \\
        SwinFuse       & 0.084 & 59.165 & 0.202 & 0.069 & 0.558 & 1.295 \\
        NestFuse           & 0.071 & 59.786 & 0.307 & 0.052 & 0.486 & 1.413 \\
        CDDFuse            & \textbf{0.065} & \textbf{60.199} & 0.469 & \textbf{\underline{0.021}} & \underline{0.845} & 1.359 \\
        DIDFuse            & \underline{0.067} & 59.988 & 0.265 & 0.062 & 0.466 & 1.459\\
        SeAFusion      & 0.094 & 58.649 & \underline{0.492} & \underline{0.044} & \textbf{\underline{0.879}} & \underline{1.472} \\
        PSFusion         & 0.067 & \underline{60.065} & 0.454 & 0.095 & 0.717 & \textbf{1.534}\\ 
        SDCFusion        & 0.078 & 59.443 & \textbf{\underline{0.534}} & \textbf{0.035} & \textbf{0.853} & 1.316\\
        CtrlFuse(Ours)   & \textbf{\underline{0.063}} & \textbf{\underline{60.317}} & \textbf{0.496} & \textbf{0.035} & 0.779 & \textbf{\underline{1.552}} \\
        \midrule
        \multicolumn{7}{c}{MSRS Dataset} \\
        \midrule
        LDFusion       & 0.056 & 61.05 & 0.438 & 0.116 & 0.541 & 1.515 \\
        SwinFuse       & 0.038 & 63.69 & 0.178 & 0.026 & 0.343 & 1.033 \\
        NestFuse       & \textbf{\underline{0.033}} & 64.128 & 0.242 & 0.025 & 0.217 & 1.138 \\
        CDDFuse        & 0.038 & \underline{64.309} & \textbf{0.689} & \underline{0.023} & \textbf{\underline{1.001}} & 1.623 \\
        DIDFuse        & \textbf{0.035} & 63.94 & 0.204 & 0.025 & 0.223 & 1.121 \\
        SeAFusion      & \underline{0.036} & \textbf{64.491} & 0.675 & \textbf{0.021} & \textbf{0.982} & 1.707 \\
        PSFusion       & 0.037 & 64.001 & 0.676 & 0.042 & 0.917 & \textbf{\underline{1.812}} \\
        SDCFusion      & 0.039 & 64.003 & \textbf{\underline{0.712}} & \underline{0.023} &0.957 & \textbf{1.739}\\
        CtrlFuse(Ours)   & \textbf{0.035} & \textbf{\underline{64.75}} & \underline{0.685} & \textbf{\underline{0.018}} & \underline{0.969} & \underline{1.726} \\
        \bottomrule
    \end{tabular}
    \caption{Quantitative comparison of CtrlFuse with 8 state-of-the-art methods. \textbf{\underline{Bold underline}} indicates the best, \textbf{Bold} indicates the second best, and \underline{Underlined} indicates the third.}
    \label{tab:comparison}
\end{table}

\subsection{Evaluation on the FMB Dataset}
\noindent \textbf{Quantitative Comparisons.}
The quantitative results on the FMB dataset are summarized in Table~\ref{tab:comparison}. Our method outperforms 8 state-of-the-art approaches on three key metrics. Achieving the best scores in PSNR and $N_{\text{abf}}$ on both the MSRS and FMB datasets demonstrates its consistent ability to preserve image clarity and reduce distortion across diverse scenarios. The top performance in $Q_{\text{abf}}$ further indicates that our method effectively retains and integrates structural and textural information from the source images. A high $N_{\text{abf}}$ score highlights strong gradient consistency and spatial coherence, which are crucial for vision-based tasks. These advantages make our method well-suited for downstream applications such as object detection.

\begin{figure}[t] 
    \centering
    \includegraphics[width=0.45\textwidth]{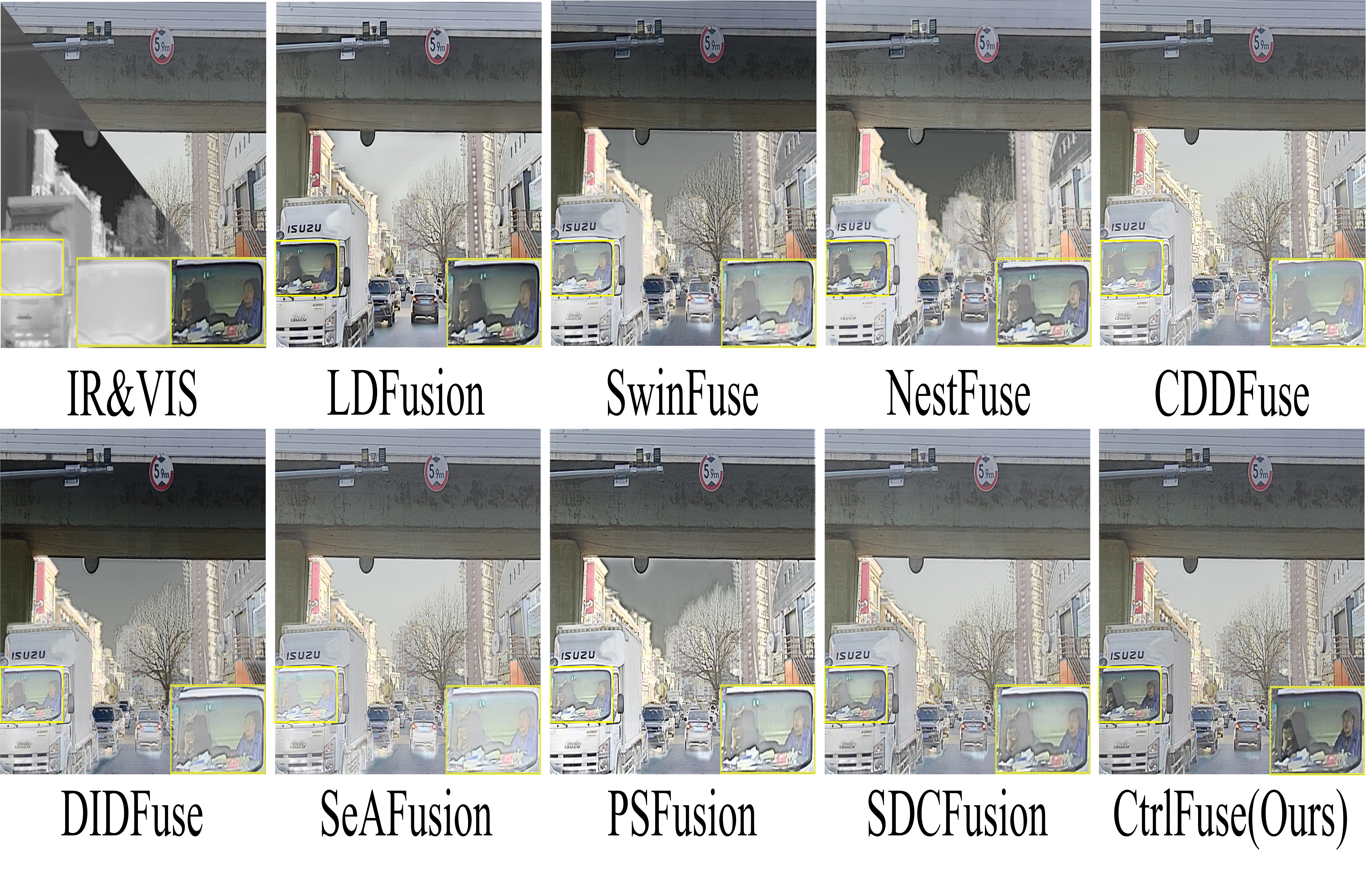} 
    \caption{Qualitative comparisons of various methods on representative images selected from the FMB dataset.} 
    \label{fig:FMBComparison}
\end{figure}

\begin{figure}[h] 
    \centering
    \includegraphics[width=0.45\textwidth]{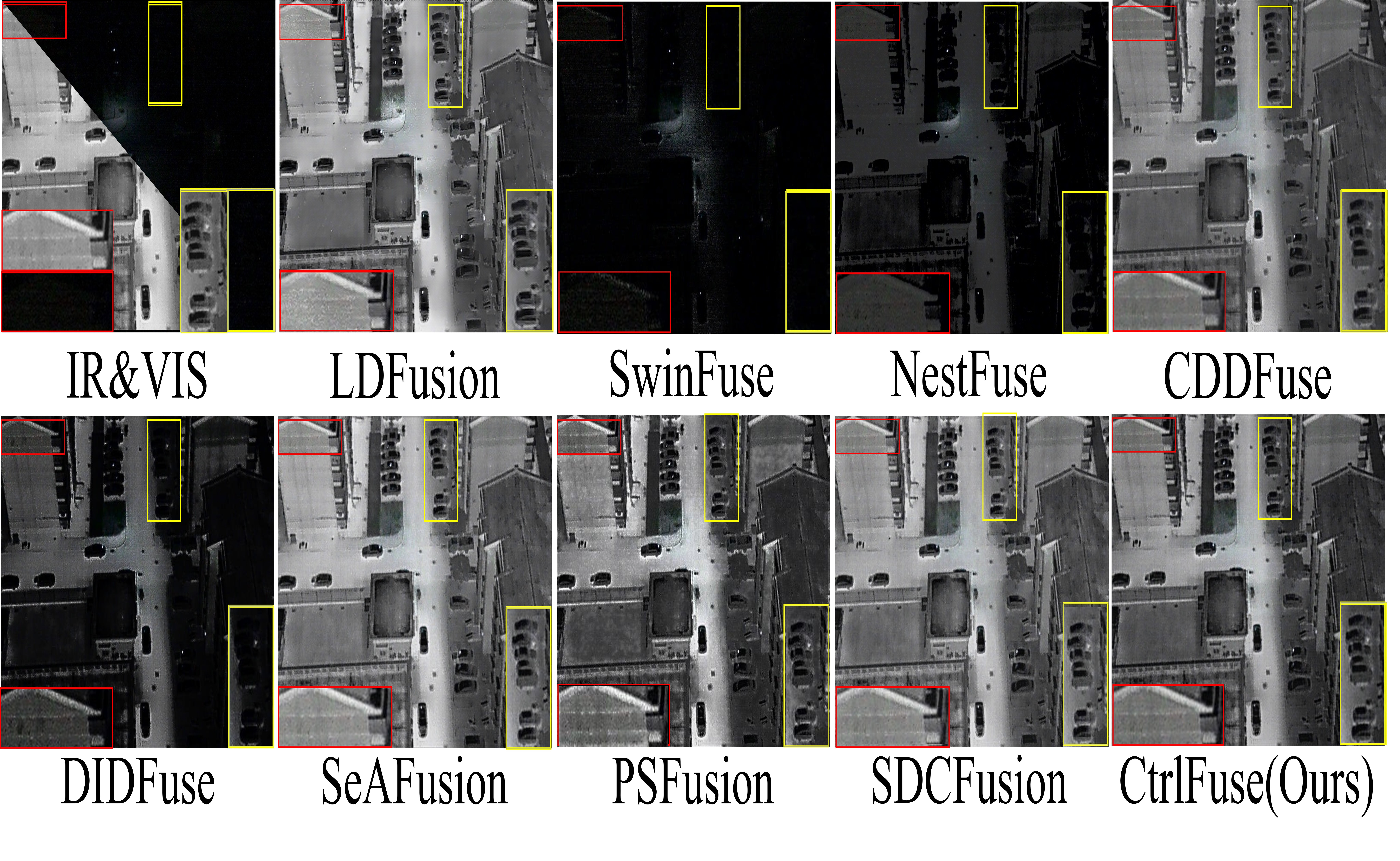} 
    \caption{Qualitative comparisons of various methods on representative images selected from DroneVehicle dataset.} 
    \label{fig:DVComparison}
\end{figure}

\begin{figure}[h] 
    \centering
    \includegraphics[width=0.45\textwidth]{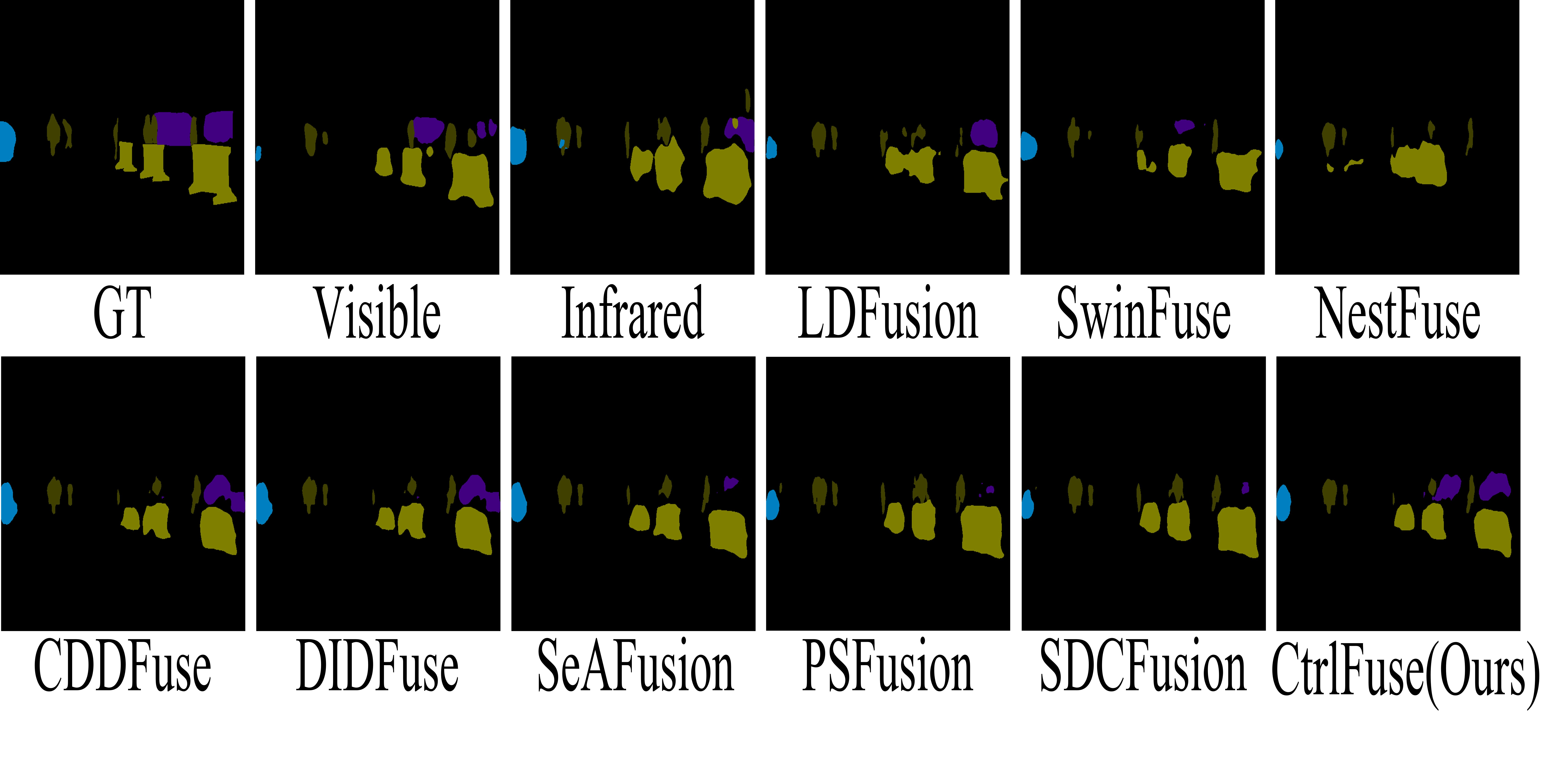} 
    \caption{Segmentation results for infrared, visible, and fused images from the MSRS dataset. The segmentation models are retrained.} 
    \label{fig:7}
\end{figure}

\begin{figure}[h]
    \centering
    \includegraphics[width=0.45\textwidth]{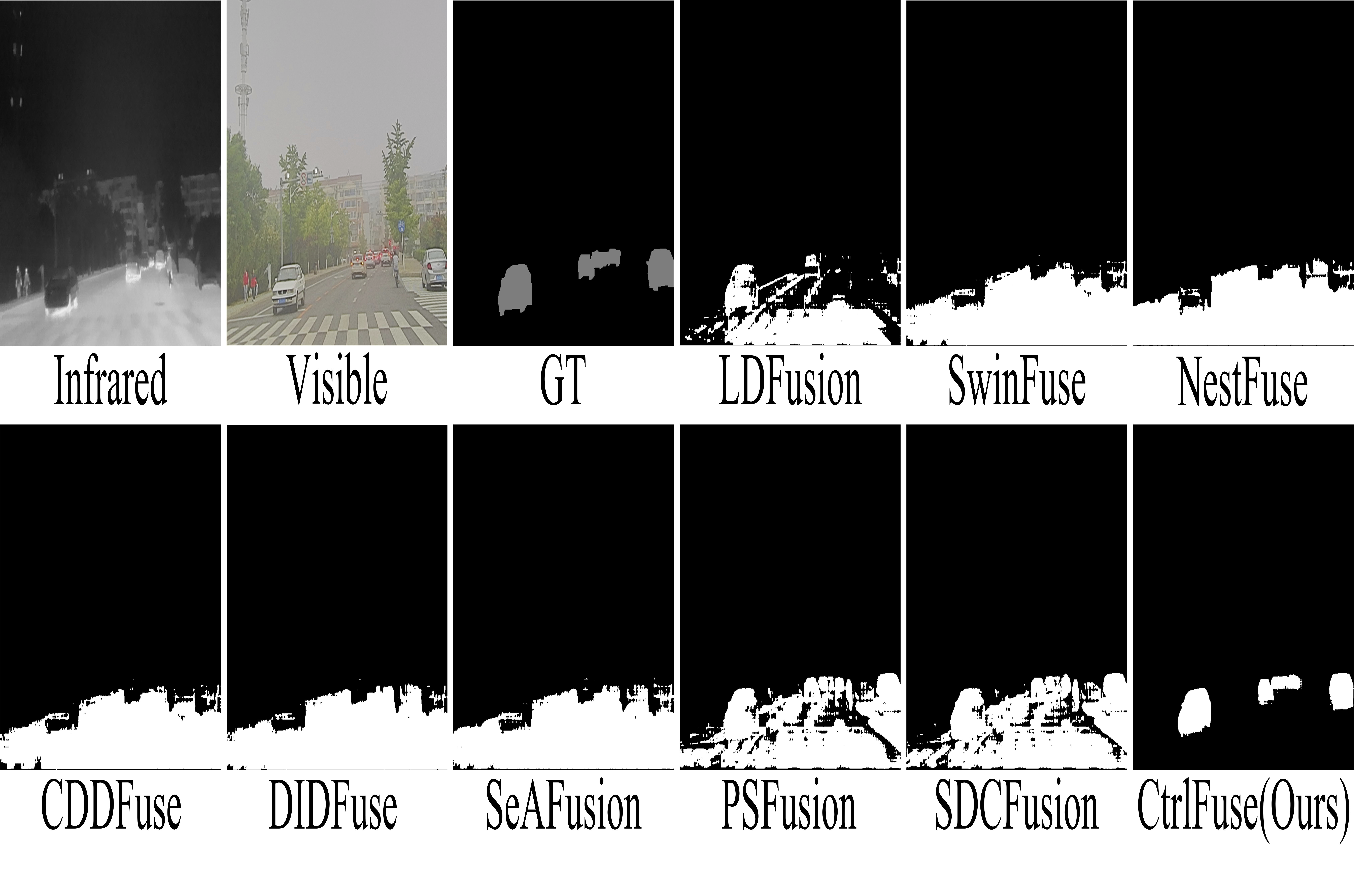} 
    \caption{Segmentation results of the original SAM model and CtrlFuse on the FMB dataset.}
    \label{fig:10Comparison}
\end{figure}

\noindent \textbf{Qualitative Comparisons.}
We mark the foreground region with the yellow rectangular box showing their zoomed-in effects for easier comparison in Fig.~\ref{fig:FMBComparison}. Among all methods except LDFusion and our proposed method, the fused images suffer from excessive brightness information in the truck's windshield due to over-enhanced infrared information, resulting in blurry appearances. Only in the fused images produced by LDFusion and our method can the person inside the truck be clearly observed. Furthermore, our method better highlights the person, making them more distinguishable from the background.

\subsection{Evaluation on the DroneVehicle Dataset}
\noindent \textbf{Quantitative Comparisons.}
Table \ref{tab:comparison} reports the performance of the different methods on the DroneVehicle dataset for 6 metrics, where our method achieves the best in 3 metrics . Among them, PSNR and MSE indicate that our method introduces minimal distortion and produces clearer and more detailed images with less noise interference. Moreover, the highest SCD indicates the method's effectiveness in maintaining sharpness and clarity. These quantitative results indicate that the proposed CtrlFuse method can efficiently capture and integrate multi-modal information, leading to fusion outcomes that are both visually compelling and rich in detail.

\noindent \textbf{Qualitative Comparisons.}
We mark the background regions with yellow and red boxes, respectively, in Fig.~\ref{fig:DVComparison}, with zoomed-in views provided for clearer comparison.As indicated by the yellow box, in the fused images of SwinFuse, NestFuse, and DIDFuse, excessive preference for the visible image makes the scene overly dark, rendering the cars in shadowed areas indistinguishable. As highlighted by the red box, LDFusion fails to adequately preserve texture details, causing the stripes on the exterior wall to be smoothed out, while SwinFuse and NestFuse still fail to distinguish the details due to insufficient infrared information. Our fused image achieves a superior balance between the infrared and visible input images.

\begin{table*}[t]
    \centering
    \setlength{\tabcolsep}{1mm} 
    \begin{tabular}{lccccccccccc}
        \toprule
        MSRS & Background & Car & Person & Bike & Curve & Car Stop & Guardrail & Color Tone & Bump & mIoU \\
        \midrule
        Infrared & 0.9669 & 0.7542 & \underline{0.5251} & 0.7287 & 0.5926 & 0.7381 & 0.9763 & 0.7925 & 0.9299 & 0.7783 \\ 
        Visible & 0.9658 & 0.7612 & 0.3904 & 0.7393 & 0.5664 & 0.7957 & 0.9797 & 0.7964 & 0.9414 & 0.7707 \\ 
        LDFusion & 0.9667 & 0.6859 & 0.4463 & 0.7221 & 0.5078 & 0.7783 & 0.9837 & 0.7867 & 0.9404 & 0.7576 \\ 
        SwinFuse & 0.9572 & 0.7013 & 0.4243 & 0.7090 & 0.5738 & 0.7757 & 0.9808 & 0.7561 & 0.9320 & 0.7567 \\
        NestFuse & 0.9587 & 0.6844 & 0.4136 & 0.6938 & 0.5731 & 0.7758 & 0.9714 & 0.7434 & 0.9349 & 0.7499 \\
        CDDFuse & 0.9696 & 0.7609 & 0.4856 & 0.7424 & 0.5843 & 0.7833 & 0.9829 & 0.7951 & 0.9375 & 0.7824 \\ 
        DIDFuse & 0.9666 & 0.7471 & 0.4625 & 0.7268 & 0.5651 & 0.8009 & 0.9817 & 0.7807 & 0.9334 & 0.7739 \\ 
        SeAFusion & 0.9695 & 0.7667 & 0.4886 & \textbf{0.7507} & 0.5991 & 0.7771 & 0.9822 & 0.7949 & 0.9395 & 0.7854 \\ 
        PSFusion & \underline{0.9708} & \underline{0.7722} & \textbf{0.5276} & \underline{0.7482} & \underline{0.6122} & 0.7962 & \underline{0.9812} & 0.8064 & \textbf{0.9443} & \underline{0.7955} \\
        SDCFusion & \textbf{0.9708} & 0.7682 & 0.5227 & 0.7420 & 0.6112 & \textbf{0.8028} & 0.9809 & \textbf{0.8121} & \underline{0.9441} & 0.7950 \\
        CtrlFuse(Ours) & 0.9705 & \textbf{0.7810} & 0.5134 & 0.7443 & \textbf{0.6179} & \underline{0.8025} & \textbf{0.9851} & \underline{0.8082} & 0.9438 & \textbf{0.7963} \\ 
        \bottomrule
    \end{tabular}  
    \caption{Segmentation performance (mIoU) of visible, infrared and fused images on the MSRS dataset. \textbf{Bold} indicates the best, \underline{Underlined} indicates the second best.}
    \label{tab:2}
\end{table*}

\subsection{High-level Vision Tasks Evaluation}
Infrared and visible image fusion integrates information from different spectral bands to produce more informative and comprehensive representations, commonly used in high-level vision tasks like object detection, classification, and scene understanding. In this section, we conduct experiments on semantic segmentation and object detection.

\noindent \textbf{Segmentation Performance.}
We conducted quantitative experiments on the MSRS dataset. Please refer to the supplementary material for detailed experimental settings. Segmentation performance, reported in Table~\ref{tab:2}, is evaluated using pixel-wise IoU. As shown, our method achieves the best performance on four categories and ranks second on four others, with the highest overall mIoU, indicating superior segmentation accuracy and generalization. We also provide visual results in Fig.~\ref{fig:7}. For the ``car" class (purple), only our method achieved complete segmentation, further demonstrating its effectiveness in preserving critical semantic information for downstream tasks. We also evaluated DeepLabV3+~\cite{chen2018encoder}, pre-trained on Cityscapes~\cite{cordts2016cityscapes}, on the FMB dataset, including the original dual-modal images, results from 8 SOTA methods, and our fused images. More details are provided in the supplementary material.

\noindent \textbf{Detection Performance.}
Object detection is a fundamental and high-level task in computer vision that aims to identify and localize multiple objects within an image or video. To adapt the model to the DroneVehicle dataset, we selected the first 2000 images from the test set. Of these, 200 were randomly chosen for evaluation, and the remaining 1800 were used to fine-tune YOLOv5 on infrared-visible image pairs, improving its detection performance on this dataset.The fine-tuned YOLOv5 model was then used to evaluate fusion methods for object detection performance.
The evaluation results are presented in Table~\ref{tab:dronevehicle_ap_5095}, which shows the average performance over thresholds from 0.5 to 0.95 (AP@[0.5:0.95]). As can be observed, the proposed method achieves the best performance in the ``car" and ``bus" categories, ranks second in the remaining categories, and maintains the highest overall metric (All), demonstrating its superior object detection accuracy across multiple evaluation criteria. 
In addition, we provide visual comparison results in the supplementary materials. We also provide a series of comparative experiments on downstream tasks without prompt masks in the supplementary materials.

\begin{table}[t]
    \centering
    \setlength{\tabcolsep}{1mm}
    \begin{tabular}{lccccc}
        \toprule
        Method & car & truck & bus & freight car & All \\
        \midrule
        LDFusion & 0.632 & 0.409 & 0.475 & 0.324 & 0.460 \\
        SwinFuse & 0.382 & 0.032 & 0.426 & 0.19 & 0.258 \\
        NestFuse & 0.555 & 0.271 & 0.365 & 0.317 & 0.377 \\
        CDDFuse & 0.632 & 0.455 & 0.470 & 0.387 & 0.486 \\
        DIDFuse & 0.528 & 0.279 & 0.388 & 0.264 & 0.364 \\
        SeAFusion & 0.646 & 0.458 & \underline{0.514} & \textbf{0.415} & 0.508 \\
        PSFusion & 0.628 & 0.475 & 0.424 & 0.277 & 0.451 \\
        SDCFusion & \underline{0.647} & \textbf{0.526} & 0.468 & 0.397 & \underline{0.510} \\
        CtrlFuse(Ours) & \textbf{0.651} & \underline{0.520} & \textbf{0.521} & \underline{0.409} & \textbf{0.525} \\
        \bottomrule
    \end{tabular}
    \caption{Object detection performance (AP@[0.5:0.95]) on the DroneVehicle dataset. \textbf{Bold} indicates the best, \underline{Underlined} indicates the second best.}
    \label{tab:dronevehicle_ap_5095}
\end{table}

\subsection{Ablation Study}
We conducted ablation studies on the MSRS dataset and reported the results in Table~\ref{tab:5}.

\begin{table}[t]
    \centering
    \setlength{\tabcolsep}{1mm}
    \begin{tabular}{lcccccc}
        \toprule
        Ablation & MSE & PSNR & $Q_{\text{abf}}$ & $N_{\text{abf}}$ & SSIM & SCD \\
        \midrule
        w/o Prompt & 0.035 & 64.958 & 0.637 & \textbf{0.014} & 0.933 & 1.635 \\
        w/o Seg & 0.036 & 64.615 & \underline{0.671} & 0.021 & \underline{0.939} & 1.636 \\
        w/o Vis & \textbf{0.033} & \textbf{65.064} & 0.656 & 0.02 & 0.915 & \underline{1.681} \\
        w/o Ir & \underline{0.034} & \underline{65.027} & 0.67 & 0.019 & 0.938 & 1.622 \\
        Exchange SQ & \underline{0.034} & 64.917 & 0.661 & 0.021 & 0.924 & 1.659 \\
        CtrlFuse(Ours) & 0.035 & 64.75 & \textbf{0.685} & \underline{0.018} & \textbf{0.969} & \textbf{1.726} \\
        \bottomrule
    \end{tabular}
    \caption{Ablation study on the MSRS dataset. \textbf{Bold} indicates the best, \underline{Underlined} indicates the second best.}
    \label{tab:5}
\end{table}

\noindent\textbf{w/o Prompt.}
To investigate the impact of prompts on fusion quality, we remove the prompt mask from our framework, retaining only the two image encoders and a single decoder. The results in Table~\ref{tab:5} show that, while the pixel-level fidelity remains similar, the perceptual quality and structural integrity are notably improved. This indicates that the prompt mask enhances the suitability of fused images for high-level vision tasks, with improvements in structural similarity, noise reduction, and spectral consistency likely benefiting downstream tasks such as object detection and semantic segmentation.

\noindent\textbf{w/o Seg.}
To verify that segmentation does not harm fusion performance, we remove the SAM mask decoder and the corresponding segmentation loss $\mathcal{L}_{\text{seg}}$. The resulting performance drop across all metrics demonstrates that segmentation improves fusion quality by preserving details, reducing noise, and maintaining structural and spectral integrity. Thus, segmentation plays a beneficial and essential role in achieving high-quality image fusion.

\noindent\textbf{w/o Vis.}
To verify the contribution of both modalities, we remove the reference prompt encoder and prompt-semantic fusion module from the visible branch, eliminating the semantic prompt $F_{\text{vis}}^{\mathrm{p}}$ in fusion. While pixel-level fidelity slightly improves, the removal degrades perceptual quality and structural integrity, indicating that the visible prompt branch plays a key role in enhancing fusion performance.

\noindent\textbf{w/o Ir.}
We perform the same ablation on the infrared branch by removing its corresponding components. A similar degradation in perceptual quality and structural integrity was observed, confirming that both modalities are crucial for high-quality image fusion.

\noindent\textbf{Exchange SQ.}
To determine whether using $F_{\text{ir}}$ or $F_{\text{vis}}$ as the support feature is more effective, we swap the roles of support and query features in the reference prompt encoder. In the modified version, $F_{\text{ref}}$ serves as the support feature, while $F_{\text{ir}}$ or $F_{\text{vis}}$ becomes the query. The overall drop in performance indicates that the original design is more effective for feature alignment and fusion.

\begin{figure}[t] 
    \centering
    \includegraphics[width=0.45\textwidth]{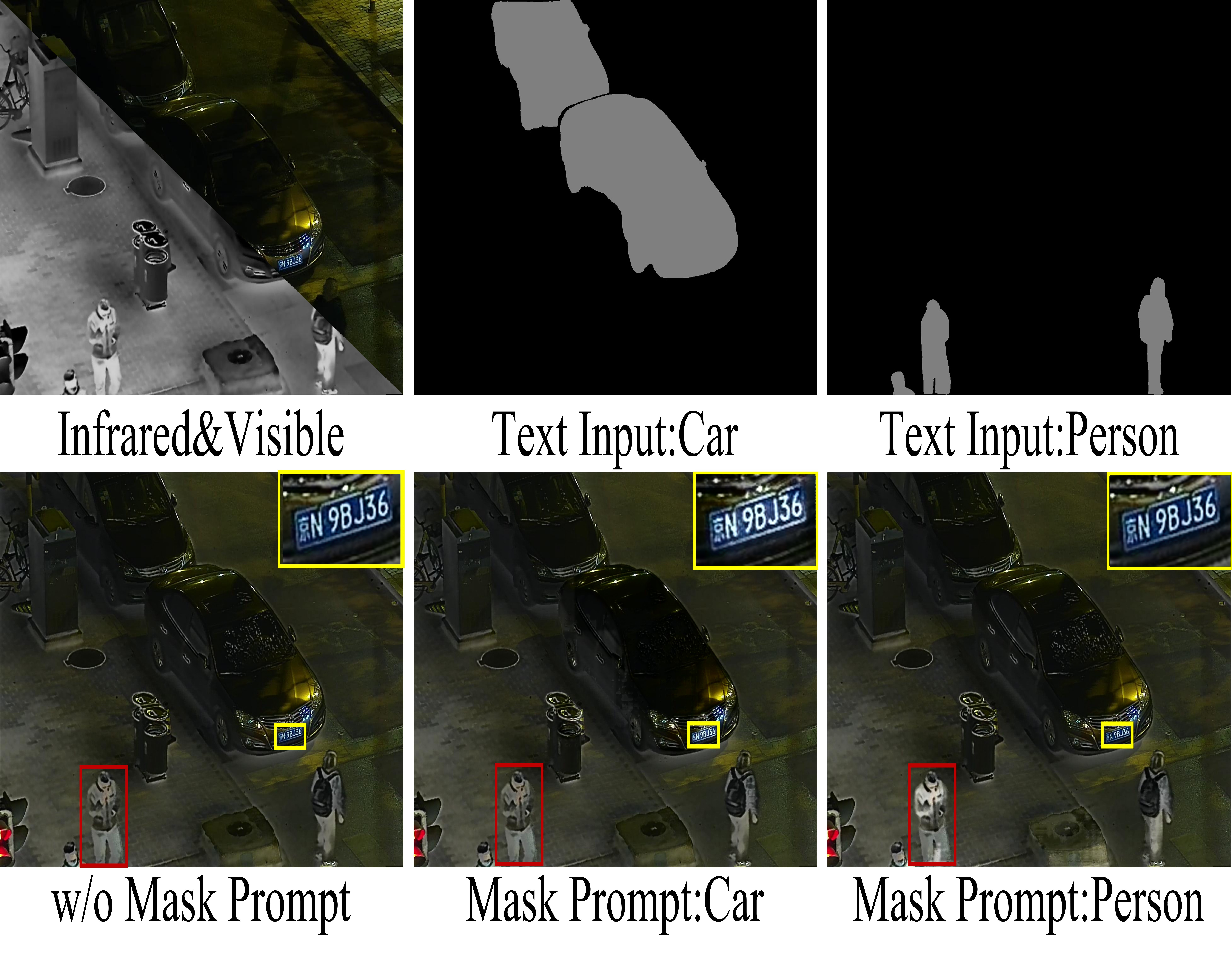} 
    \caption{Fusion results of CtrlFuse with different prompt masks generated by Grounded-SAM on the LLVIP dataset.} 
    \label{fig:LLVIPControl}
\end{figure}

\subsection{Analysis and Discussion}
\noindent\textbf{The Impact of Mask Prompt Fine-tuning on SAM Performance.}
To investigate how the fusion branch enhances segmentation via prompt fine-tuning, we compared CtrlFuse with the original SAM and other fusion methods. As shown in Fig.~\ref{fig:10Comparison}, our method, which directly combines the segmentation masks from its two branches, achieves superior performance, demonstrating that the fine-tuned CtrlFuse enables more effective personalized segmentation guided by mask prompts.

\noindent\textbf{Controllability of CtrlFuse.}
CtrlFuse uses prompt masks to guide the model to focus on and enhance specific targets. To validate this, we experimented on the LLVIP dataset~\cite{jia2021llvip}, which lacks segmentation annotations. We employed Grounded-SAM~\cite{ren2024grounded} to generate masks from different text inputs and used them to guide the fusion process. The resulting images in Fig.~\ref{fig:LLVIPControl} demonstrate the model's capability to emphasize designated classes under varying conditions.

\noindent\textbf{Sensitivity Analysis on Prompt Mask.}
\begin{figure}[t] 
    \centering
    \includegraphics[width=0.45\textwidth]{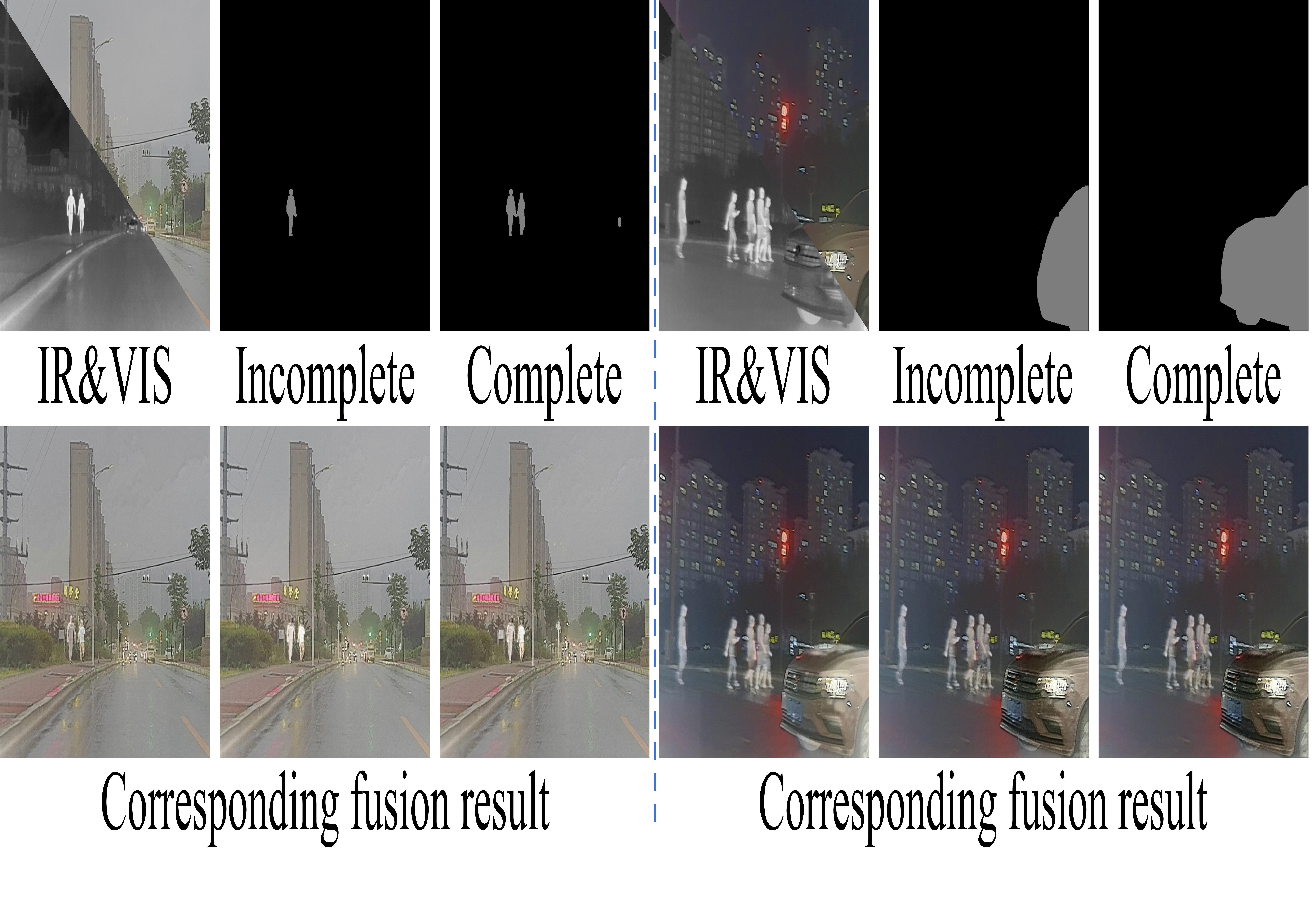} 
    \caption{Comparison of fusion results under complete versus incomplete or coarse prompt masks.} 
    \label{fig:sensitivity}
\end{figure}
Our analysis shows the model's robustness to prompt mask quality. Even when masks are incomplete or of low quality (e.g., masking only one object or just a part of it), the fusion results still effectively highlight the target objects, including unmarked areas, as shown in Fig.~\ref{fig:sensitivity}. Visual comparisons confirm that mask quality has no significant impact on the output.

\section{Conclusion}
In this paper, we present CtrlFuse, a mask-prompt controllable multimodal image fusion framework that establishes dynamic interactions between scene understanding requirements and low-level fusion processes. By integrating mask-guided semantic adaptation with explicit feature fusion mechanisms, the proposed method addresses the critical limitation of conventional task-driven fusion approaches in rigid semantic constraints. Through synergistic design of the reference prompt encoder for adaptive semantic tuning, the prompt semantic fusion module for cross-modal feature aggregation, and the joint optimization of segmentation-aware fusion objectives, our framework achieves both semantically enhanced fusion results and improved downstream perception accuracy. Experimental validations across diverse scenarios demonstrate that the explicit injection of semantic guidance enables not only controllable fusion behavior but also mutual reinforcement between multimodal fusion and semantic segmentation tasks. 
The proposed paradigm provides an interactive and controllable multimodal fusion perception solution for intelligent unmanned systems, which is particularly suitable for all-weather search and rescue applications that require a focus on specific targets.

\section{Acknowledgments}
This work was supported by the National Natural Science Foundation of China under Grants 62222608, 62506073, and 62436002, the Tianjin Natural Science Funds for Distinguished Young Scholar under Grant 23JCJQJC00270, the Postdoctoral Fellowship Program of CPSF under Grant Number GZB20250395, the Jiangsu Funding Program for Excellent Postdoctoral Talent under Grant Number 2025ZB294, the the Zhejiang Provincial Natural Science Foundation of China under Grant LD24F020004.

\bibliography{aaai2026}

\newpage
\section{Appendices.}

\begin{figure*}[t]
	\centering
	\includegraphics[width=1.0\textwidth]{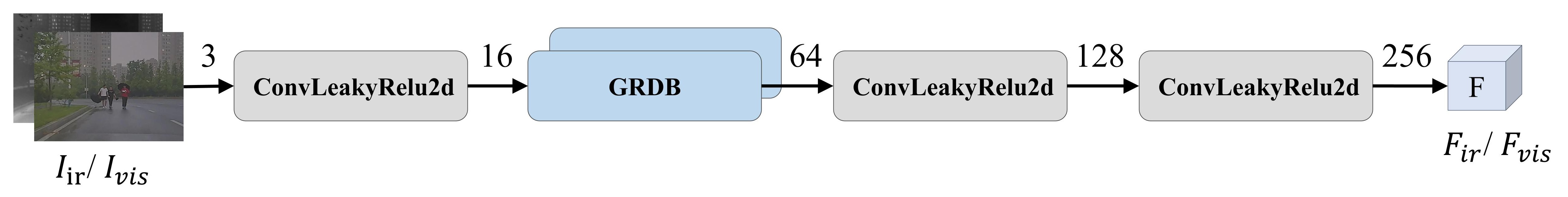} 
	\caption{The architecture of CtrlFuse Image Encoder. Encoder consists of GRDB, convolutional layers, and activation layers.}
	\label{fig:enc}
\end{figure*}

\begin{figure*}[t] 
    \centering
    \includegraphics[width=1.0\textwidth]{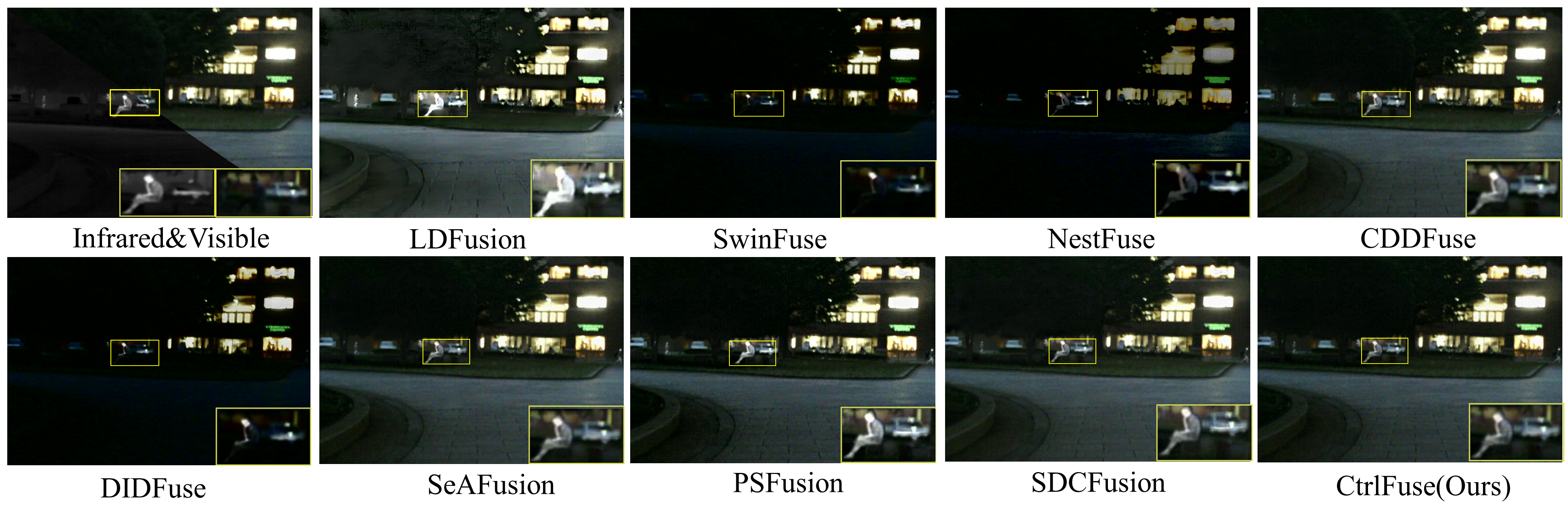} 
    \caption{Qualitative comparisons of various methods on representative images selected from the MSRS dataset.} 
    \label{fig:MSRSComparison}
\end{figure*}

\subsection{Fusion Evaluation Metric}

\noindent \textbf{MSE.} The Mean Squared Error (MSE) is a commonly used image quality evaluation metric. It calculates the average of the squared differences between the corresponding pixel values of the original and processed images. Assuming the original image $I$ and the fused image $J$ both have a size of $M\times N$, the Mean Squared Error (MSE) can be defined as:
\begin{equation}
MSE=\frac{1}{MN} \sum{ }_{i=1}^M \sum{ }_{j=1}^N[I(i, j)-J(i, j)]^2.
\end{equation}
The lower the MSE, the better the fused image is in terms of similarity to the source images. A lower MSE indicates less distortion and more accurate preservation of details and information from the source images.

\noindent \textbf{PSNR.} The Peak Signal-to-Noise Ratio (PSNR) is a widely used image quality evaluation metric. It is based on the Mean Squared Error (MSE) and reflects the ratio between the maximum possible signal intensity and the noise intensity in an image.
\begin{equation}
PSNR=10 \log _{10}\left(\frac{MAX^2}{MSE}\right),
\end{equation}
where $MAX^2$ is the square of the maximum pixel value in the fused image.
The higher the PSNR, the more effectively the fused image retains the quality and integrity of the source images. A high PSNR suggests superior visual fidelity, indicating successful preservation of original details and minimal loss or distortion during the fusion process.

\noindent \textbf{$Q_{\text{abf}}$.} $Q_{abf}$ refers to a specific image quality evaluation metric, commonly used to measure the similarity and fusion effectiveness between fused images and source images. It comprehensively considers multiple aspects of image information, such as brightness, contrast, and structure. By calculating similarity measures between the fused image and the source images in different feature spaces, and then combining them through weighted summation, $Q_{abf}$ produces an overall quality assessment score. It can be calculated as:
\begin{equation}
\begin{split}
Q_{abf} = \frac{1}{N} \sum_{i=1}^N w_i 
&\frac{(2 \mu_{a,i} \mu_{b,i} + C_1)}{(\mu_{a,i}^2 + \mu_{b,i}^2 + C_1)} \\
&\cdot \frac{(2 \sigma_{a,b,i} + C_2)}{(\sigma_{a,i}^2 + \sigma_{b,i}^2 + C_2)},
\end{split}
\end{equation}
where $N$ is the number of image blocks, $w_i$ is the weight of the $i$-th image block,$\mu_{a, i}$ and $\mu_{b, i}$ are the means of the $i$-th image block in source images $a$ and $b$, respectively,$\sigma_{a, i}^2$ and $\sigma_{b, i}^2$ are the variances of the $i$-th image block in source images $a$ and $b$, respectively,$\sigma_{a, b, i}$ is the covariance of the $i$-th image block in source images a and b, and$C_1$ and $C_2$ are constants. A higher $Q_{abf}$ indicates more effective information transfer and better visual performance of the fused image, reflecting a well-balanced fusion outcome with minimal degradation.

\noindent \textbf{$N_{\text{abf}}$.} $N_{abf}$ correlates with factors including the image's noise level and the extent of detail retention. It assesses image quality by analyzing and quantifying characteristics such as noise properties and high-frequency detail information within the image.
\begin{equation}
N_{a b f}=\frac{1}{N} \sum_{i=1}^N w_i \frac{\sigma_{n, i}}{\sigma_{a, b, i}+C},
\end{equation}
where $N$ is the number of image blocks, $w_i$ is the weight of the $i$-th image block, $\sigma_{n, i}$ is the standard deviation of the noise in the $i$-th image block, $\sigma_{a, b, i}$ is the covariance of the $i$-th image block in source images a and b, and $C$ is a small constant used to prevent division by zero.
A lower $N_{abf}$ reflects a higher-quality fused image with less noise and fewer artifacts, making it clearer and more faithful to the source images.

\noindent \textbf{SCD.} The Sum of Correlation Differences is a metric used to measure the differences between fused images and source images as a way to characterize the quality of the fusion algorithm. 
\begin{equation}
\mathrm{SCD}=\sum_{i=1}^N w_i\left|\rho_{a, i}+\rho_{b, i}-2 \rho_{f, i}\right|,
\end{equation}
where $N$ is the number of image blocks. $w_i$ is the weight of the $i$-th image block.$\rho_{a, i}$ and $\rho_{b, i}$ are the correlation coefficients of the $i$-th block in source images $a$ and $b$, respectively. $\rho_{f, i}$ is the correlation coefficient of the $i$-th block in the fused image. A higher SCD indicates that the fused image contains richer information from the source images.

\noindent \textbf{SSIM.} SSIM evaluates image quality based on three components: luminance (mean intensity), contrast (variance), and structural similarity (correlation). It is defined as:
\begin{equation}
\operatorname{SSIM}(x, y)=\frac{\left(2 \mu_x \mu_y+C_1\right)\left(2 \sigma_{x y}+C_2\right)}{\left(\mu_x^2+\mu_y^2+C_1\right)\left(\sigma_x^2+\sigma_y^2+C_2\right)},
\end{equation}
where $\mu_x$ and $\mu_y$ are the averages of the pixel values in the two images (representing luminance). $\sigma_x^2$ and $\sigma_y^2$ are the variances of the pixel values in the two images (representing contrast). $\sigma_{xy}$ is the covariance between the two images (representing structural similarity or correlation). $C_1$ and $C_2$ are small constants added to stabilize the division with a weak denominator.
An increase in SSIM suggests that the fused image better captures the structural characteristics of the source images, leading to a result that appears more visually consistent and natural to human observers.

\begin{figure*}[t]
	\centering
	\includegraphics[width=1.0\textwidth]{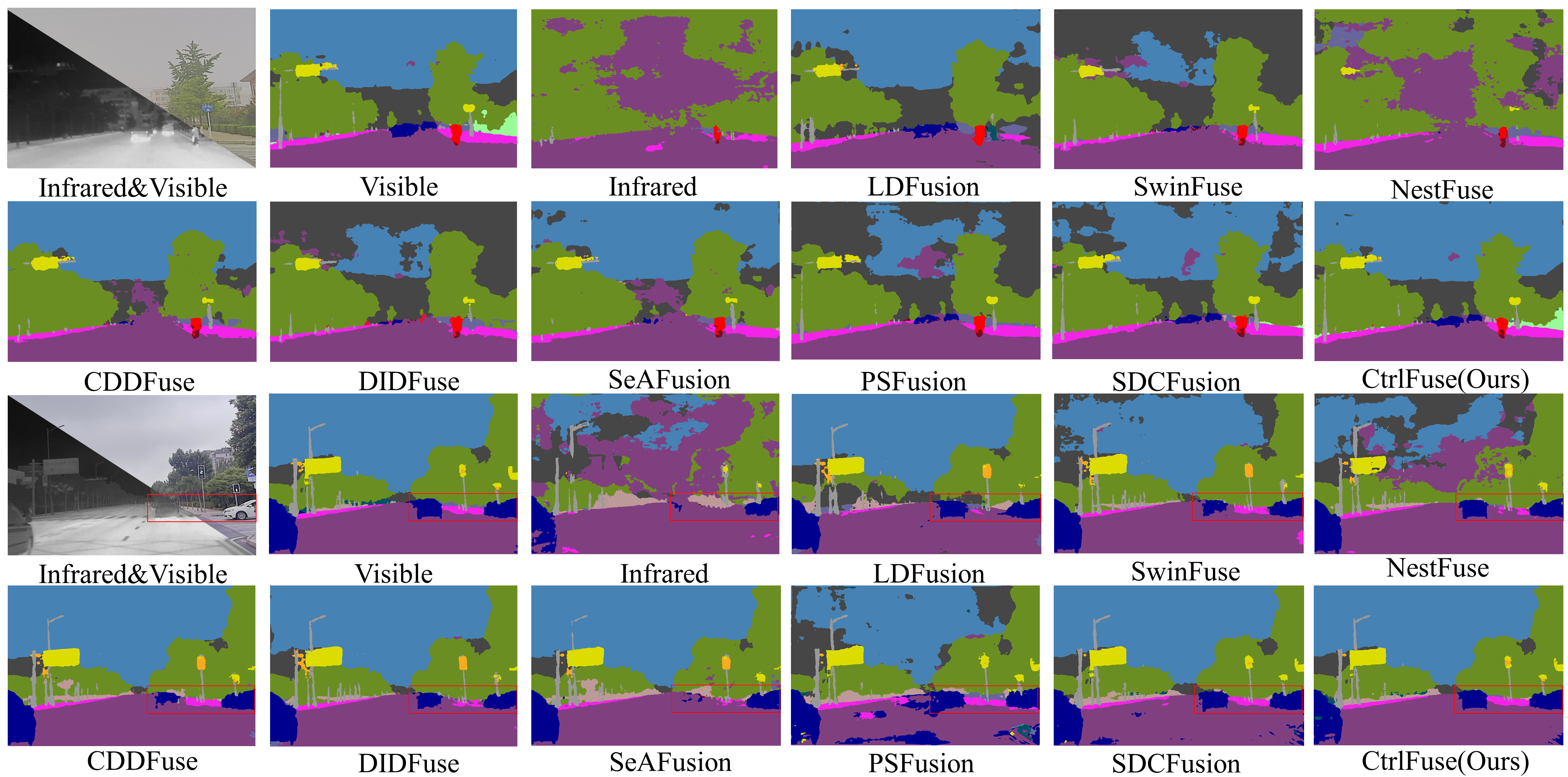} 
	\caption{Segmentation results for infrared, visible, and fused images from the FMB dataset. The segmentation model is Deeplabv3+, pre-trained on the Cityscapes dataset.}
	\label{fig:presegComparison}
\end{figure*}

\subsection{Model Details}
In the CtrlFuse method, both the infrared and visible image encoders utilize Gradient Residual Dense Blocks (GRDB) proposed in ~\cite{tang2022image}. In addition to the Gradient Residual Dense Blocks (GRDB), the encoders also consist of convolutional layers and activation layers. The detailed structure of the image encoders is shown in Fig.~\ref{fig:enc}.The decoder module is implemented as a sequence of ConvLeakyRelu2d layers. The channel dimensions are successively halved (aside from the final step) through the network, following the pattern: 256 → 128 → 64 → 32 → 16 → 1, to ultimately generate a single-channel output.

\begin{table}[t]
\centering
\setlength{\tabcolsep}{1mm}
\begin{tabular}{lcccccc}
\toprule
Methods & 
\shortstack{MSE} & 
\shortstack{PSNR} & 
\shortstack{$Q_{abf}$} & 
\shortstack{$N_{abf}$} & 
\shortstack{SSIM} & 
\shortstack{SCD} \\
\midrule
LDFusion       & 0.056 & 61.05 & 0.438 & 0.116 & 0.541 & 1.515 \\
SwinFuse       & 0.038 & 63.69 & 0.178 & 0.026 & 0.343 & 1.033 \\
NestFuse       & \textbf{\underline{0.033}} & 64.128 & 0.242 & 0.025 & 0.217 & 1.138 \\
CDDFuse        & 0.038 & \underline{64.309} & \textbf{0.689} & \underline{0.023} & \textbf{\underline{1.001}} & 1.623 \\
DIDFuse        & \textbf{0.035} & 63.94 & 0.204 & 0.025 & 0.223 & 1.121 \\
SeAFusion      & \underline{0.036} & \textbf{64.491} & 0.675 & \textbf{0.021} & \textbf{0.982} & 1.707 \\
PSFusion       & 0.037 & 64.001 & 0.676 & 0.042 & 0.917 & \textbf{\underline{1.812}} \\
SCDFusion      & 0.039 & 64.003 & \textbf{\underline{0.712}} & \underline{0.023} & 0.957 & \textbf{1.739} \\
CtrlFuse(Ours) & \textbf{0.035} & \textbf{\underline{64.75}} & \underline{0.685} & \textbf{\underline{0.018}} & \underline{0.969} & \underline{1.726} \\
\bottomrule
\end{tabular}
\caption{Quantitative comparison of our CtrlFuse with 8 state-of-the-art methods on the MSRS dataset. \textbf{\underline{Bold underline}} indicates the best performance, \textbf{bold} the second best, and \underline{underline} the third best.}
\label{tab:msrs_comparison}
\end{table}

\begin{figure}[!htbp]
\centering
\begin{tikzpicture}
\begin{axis}[
    width=0.75\linewidth,
    height=0.45\linewidth,  
    xlabel={$N$},
    ylabel={mIoU},
    xmin=1, xmax=5,
    ymin=0.700, ymax=0.726,
    xtick={1,2,3,4,5},
    ytick={0.700,0.705,0.710,0.715,0.720,0.725},
    grid=major,
    major grid style={line width=0.2pt, draw=gray!25},
    axis lines=left,
    tick label style={font=\footnotesize}
]

\addplot[blue, mark=*, semithick, mark size=1.8pt] 
coordinates {
    (1,0.7158)
    (2,0.7183) 
    (3,0.7172)
    (4,0.7237)
    (5,0.7028)
};
\end{axis}
\end{tikzpicture}
\caption{Hyperparameter Optimization of Query Vector Count in RPE Module}
\label{fig:Q_hyperparameter}
\end{figure}
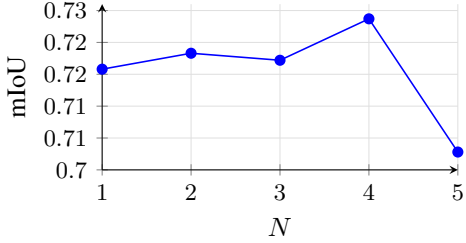

\begin{figure}[t]
	\centering
	\includegraphics[width=0.45\textwidth]{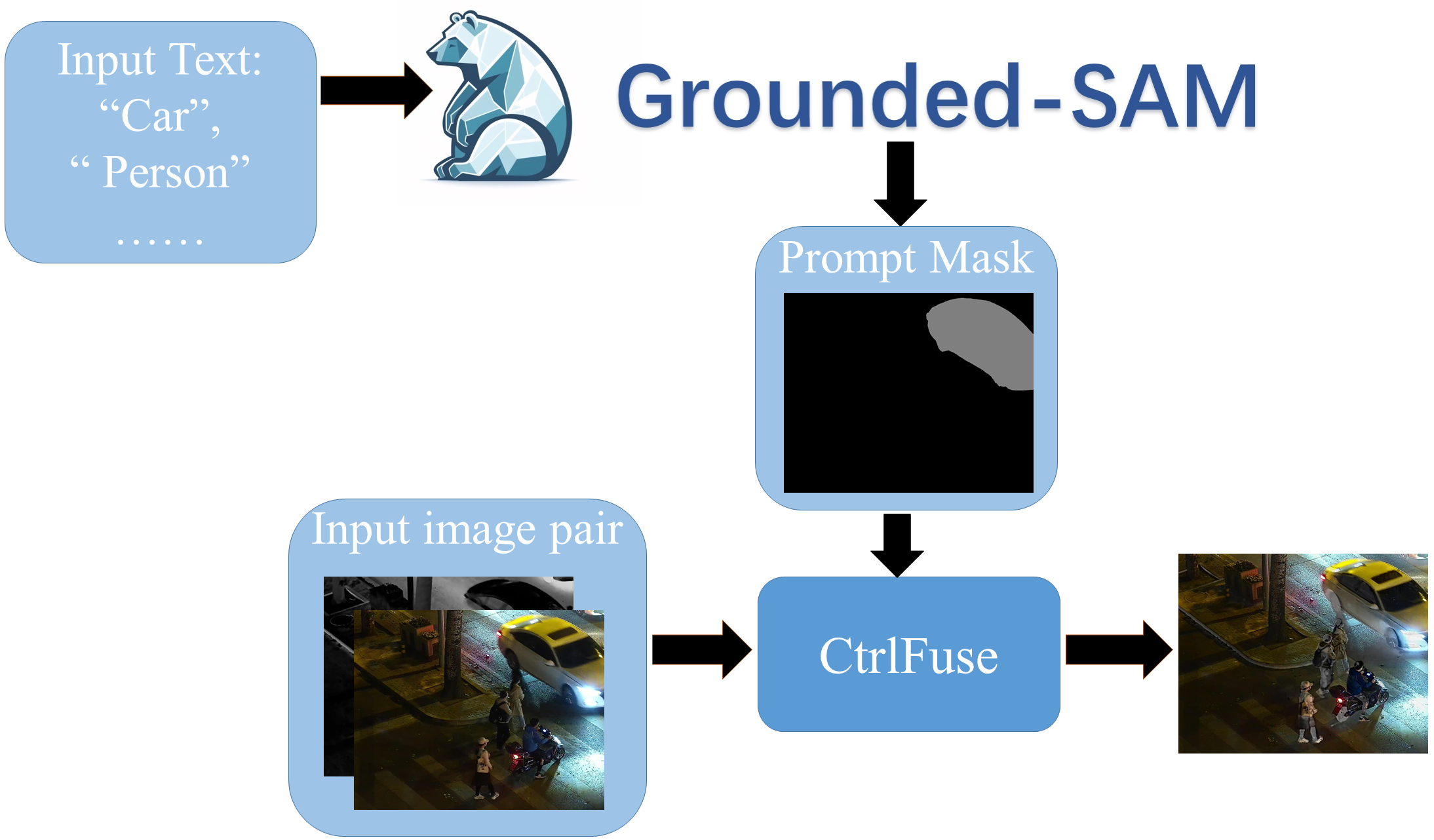} 
	\caption{Inference pipeline with Prompt Mask.}
	\label{fig:Ctrltest}
\end{figure}

\begin{figure*}[t]
	\centering
	\includegraphics[width=1.0\textwidth]{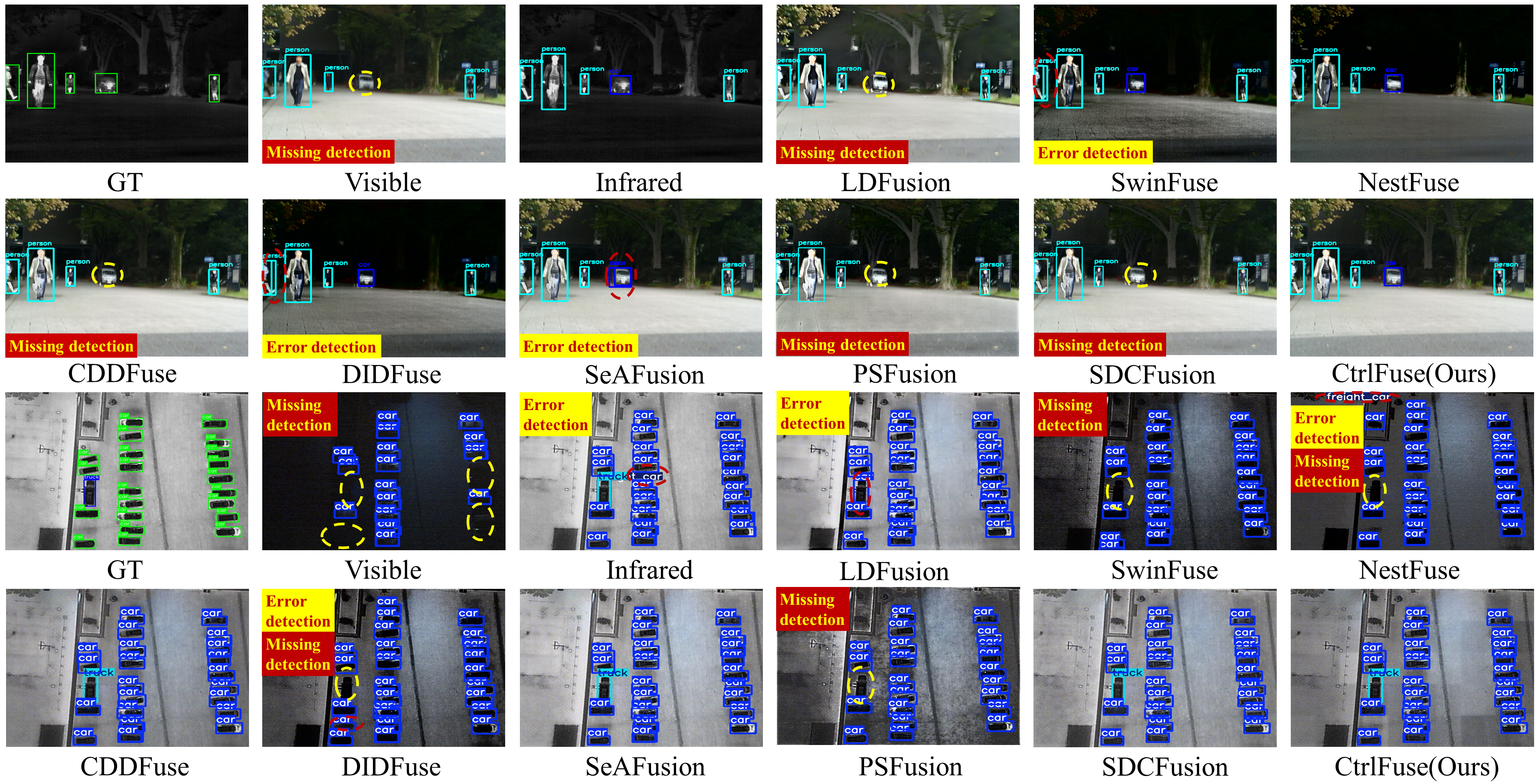} 
	\caption{Object detection results for infrared, visible, and fused images from the MSRS and
 DroneVehicle dataset.}
	\label{fig:MDdetect}
\end{figure*}

In addition, our model utilizes a frozen SAM model, specifically $\texttt{sam\_vit\_h\_4b8939}$. The image decoder is composed of convolutional layers and activation layers.

\subsection{Hyperparameter Experiment about the Learnable Query}
In the RPE module, we introduce a set of learnable query vectors $Q$ to steer the attention mechanism toward the target regions specified by the prompts. To determine the optimal number of query vectors $N$, we conduct a series of hyperparameter experiments. Given the primary role of $Q$ in refining the segmentation outputs from SAM to produce more accurate binary masks, we adopt the mIoU trend with respect to $N$ as the selection criterion. As shown in Fig.~\ref{fig:Q_hyperparameter} depicting mIoU versus $N$, the model achieves optimal performance when $N = 40$, which is consequently chosen as the final configuration.

\subsection{Loss Function}
In CtrlFuse, we use $\mathcal{L}_{\text {fusion }}$ to guide the optimization of the image fusion network, and the segmentation branch is fine-tuned through the segmentation loss $\mathcal{L}_{\text {seg}}$. The entire framework is trained end-to-end by jointly optimizing the two loss functions. 
Therefore, the overall loss is defined as:
\begin{equation}
\mathcal{L}_{\text {total}}=\mathcal{L}_{\text {fusion}}+ \mathcal{L}_{\text {seg}}.
\end{equation}
Specifically, $\mathcal{L}_{\text {fusion }}$ can be calculated as:
\begin{equation}
\mathcal{L}_{\text {fusion }}=\mathcal{L}_{\text {pixel }}\\+ \mathcal{L}_{\text {grad }}+\mathcal{L}_{\text {int }}+\mathcal{L}_{\text {percep}},
\end{equation}
where $\mathcal{L}_{\text {pixel }}$ ~\cite{sun2022detfusion} is composed of two components: one for the object regions and another for the background region. It can be described as:
\begin{equation}
\begin{split}
\mathcal{L}_{\text{pixel}} 
&= \frac{1}{H W} \left\| I_{seg} \cdot \left(I_F - \max\left(I_{vis}, I_{ir}\right)\right) \right\|_1 \\
&\quad + \frac{1}{H W} \left\| (1 - I_{seg}) \cdot \left(I_F - \text{mean}\left(I_{vis}, I_{ir}\right)\right) \right\|_1,
\end{split}
\end{equation}
where $H$ and $W$ are the height and width of an image, respectively, 
$\|\cdot\|_1$ stands for the $\ell_1$-norm, and $I_{seg}$ represents the segmentation results obtained from the two branches.The gradient loss plays a crucial role in improving the structural quality and visual fidelity of the fused image, focusing on preserving spatial gradients. We define the gradient loss of infrared and visible images as:
\begin{equation}
\mathcal{L}_{g r a d}=\frac{1}{H W}\left\|\nabla I_F-\max \left(\nabla I_{v i s},\nabla I_{i r}\right)\right\|_1.
\end{equation}

The intensity loss is used to preserve the pixel-wise brightness information and ensure visual fidelity between the fused image and the source images. Therefore, we define the intensity loss of infrared and visible images as:
\begin{equation}
\mathcal{L}_{i n t}=\frac{1}{H W}\left\|I_F-\max \left(I_{i r}, I_{vis}\right)\right\|_1.
\end{equation}

The perceptual loss enhances the visual realism and structural fidelity of fused images by aligning their deep features with those of the source images. $\mathcal{L}_{\text {percep }}$can be defined as:
\begin{equation}
\begin{split}
\mathcal{L}_{\text{percep}} = 
&\sum_{l \in L} \frac{1}{N_l} \left\| \Phi_l(I_{\text{F}}) - \Phi_l(I_{\text{ir}}) \right\|_2^2 \\
+ &\sum_{l \in L} \frac{1}{N_l} \left\| \Phi_l(I_{\text{F}}) - \Phi_l(I_{\text{vis}}) \right\|_2^2,
\end{split}
\end{equation}
where $\Phi_l(\cdot)$ represents the feature map obtained from the $l$-th layer of the pre-trained VGG model, $N_l$ is the normalization factor corresponding to the number of elements in the feature map, and $\|\cdot\|_2^2$ denotes the squared L2 norm, which is equivalent to the mean squared error.

Ablation studies on the perceptual loss confirm its effectiveness in enhancing both the visual realism and structural fidelity of the fusion results. The corresponding quantitative results are presented in Table~\ref{tab:perceptual_loss_ablation}.

\begin{table}[h]
\centering
\setlength{\tabcolsep}{1mm}
\fontsize{9}{11}\selectfont
\begin{tabular}{lcccccc}
\toprule
Method & 
\shortstack{MSE} & 
\shortstack{PSNR} & 
\shortstack{Qabf} & 
\shortstack{Nabf} & 
\shortstack{SSIM} & 
\shortstack{SCD} \\
\midrule
w/o $\mathcal{L}_{percep}$ & 0.046 & 62.922 & 0.715 & 0.033 & 0.912 & 1.434 \\
CtrlFuse (Ours) & \textbf{0.043} & \textbf{63.292} & \textbf{0.719} & \textbf{0.024} & \textbf{0.925} & \textbf{1.522} \\
\bottomrule
\end{tabular}
\caption{Ablation study on perceptual loss.}
\label{tab:perceptual_loss_ablation}
\end{table}

We employ the Binary Cross-Entropy (BCE) loss and Dice loss to jointly supervise the segmentation learning process. Therefore we can define $\mathcal{L}_{\text {seg}}$ as:
\begin{equation}
\mathcal{L}_{\text {seg }}=\mathcal{L}_{\text {BCE }}+ \mathcal{L}_{\text {dice }},
\end{equation}
\begin{equation}
\mathcal{L}_{\mathrm{BCE}}=-\frac{1}{N} \sum_{i=1}^N\left[y_i \log \left(\hat{y}_i\right)+\left(1-y_i\right) \log \left(1-\hat{y}_i\right)\right],
\end{equation}
\begin{equation}
\mathcal{L}_{\mathrm{dice}} =1 - \frac{2 \sum_{i=1}^N \left( \hat{y}_i \cdot y_i \right)}{\sum_{i=1}^N \hat{y}_i^2 + \sum_{i=1}^N y_i^2},
\end{equation} 
where $y_i$ is the ground-truth label for the $i$-th pixel, $\hat{y}_i$ is the predicted probability for the $i$-th pixel, and $N$ is the total number of pixels.

\subsection{Evaluation on the MSRS dataset}
\noindent \textbf{Quantitative Comparisons.}
Table~\ref{tab:msrs_comparison} presents the results of the quantitative evaluation on the MSRS dataset, where our method achieves the best performance in two metrics and ranks second and third in the remaining metrics, respectively. In particular, CtrlFuse attains the highest values for PSNR and $N_{abf}$, indicating its superior performance in preserving image quality and structural details. The highest PSNR value further corroborates this, as it reflects excellent pixel-level similarity. The lowest $N_{abf}$ value highlights the method’s effectiveness in minimizing blurring and maintaining high-resolution image quality. 

\noindent \textbf{Qualitative Comparisons.}
We mark the foreground region with the yellow rectangular box. We also show their zoomed-in effects for easier comparison in Fig.~\ref{fig:MSRSComparison}. In LDFusion, the over-reliance on thermal signals in foreground regions disrupts background texture preservation, causing artifacts such as blurred or distorted vehicles. Our method addresses this by adaptively suppressing redundant thermal interference while maintaining visible modality details, ensuring balanced multi-modal fusion. In SwinFuse, NestFuse, and DIDFuse, excessive contrast enhancement in thermal regions leads to over-darkening of visible details, severely weakening the saliency of target objects. This imbalance causes critical features to become indistinct due to overly suppressed luminance. 

\subsection{Ablation Study}
In this ablation study, we removed the RPE and PSFM modules, retaining only the SAM as the segmentation component. Under this configuration, SAM directly performs multi-class semantic segmentation without requiring input prompt masks. The semantic masks generated by the two branches are normalized and fused with their respective extracted features via Hadamard product, then added to the reference feature $F_{ref}$ to obtain the final fused feature. The rest of the model architecture remains unchanged. The final quantitative results are presented in Table~\ref{tab:rpe_psfm_ablation}.

\begin{table}[h]
\centering
\setlength{\tabcolsep}{1mm}
\fontsize{9}{11}\selectfont
\begin{tabular}{lcccccc}
\toprule
Method & 
\shortstack{MSE} & 
\shortstack{PSNR} & 
\shortstack{Qabf} & 
\shortstack{Nabf} & 
\shortstack{SSIM} & 
\shortstack{SCD} \\
\midrule
w/o RPE\&PSFM & 0.049 & 62.682 & 0.691 & 0.039 & 0.895 & 1.463 \\
CtrlFuse (Ours) & \textbf{0.043} & \textbf{63.292} & \textbf{0.719} & \textbf{0.024} & \textbf{0.925} & \textbf{1.522} \\
\bottomrule
\end{tabular}
\caption{Quantitative results of the ablation study on RPE and PSFM modules.}
\label{tab:rpe_psfm_ablation}
\end{table}

\begin{table*}[t]
    \centering
    \setlength{\tabcolsep}{2pt}
    \begin{tabular}{lcccccccccc}
        \toprule
        Method & Background & Car & Person & Bike & Curve & Car Stop & Guardrail & Color Tone & Bump & mIoU \\
        \midrule
        w/o Mask Prompt & 0.9704 & 0.7698 & 0.5125 & \textbf{0.7531} & 0.6156 & \textbf{0.8026} & 0.9850 & 0.8066 & \textbf{0.9447} & 0.7956 \\
        CtrlFuse(Ours) & \textbf{0.9705} & \textbf{0.7810} & \textbf{0.5134} & 0.7443 & \textbf{0.6179} & 0.8025 & \textbf{0.9851} & \textbf{0.8082} & 0.9438 & \textbf{0.7963} \\
        \bottomrule
    \end{tabular}
    \caption{Quantitative comparison of segmentation performance with and without mask prompts on the MSRS dataset.\textbf{Bold} indicates the best.}
    \label{tab:mask_prompt_segcomparison}
\end{table*}

\subsection{High-level Vision Tasks Evaluation}
\noindent \textbf{Segmentation Performance.}
We conducted quantitative experiments on the MSRS dataset. We retrained the DeepLabV3+ semantic segmentation models for both the dual-modal image pairs and nine fusion methods, including our approach, using identical parameters throughout the training process. We employ DeeplabV3+ with a ResNet backbone, setting the initial learning rate to 0.01 and utilizing a single GPU for training. The model is configured to train over 200 epochs with a batch size of 11. 

Moreover, we evaluate the performance of DeepLabV3+, pre-trained on the Cityscapes dataset, on both the original dual-modal images, the fused results generated by eight state-of-the-art fusion methods, and our proposed fused images on the FMB dataset. This experimental setup allows for a comprehensive comparison of how different fusion strategies impact downstream semantic segmentation performance.
Given that the Cityscapes dataset is composed of visible light imagery, the visible images can be used as an approximate ground truth for evaluation purposes. In Fig.~\ref{fig:presegComparison}, focusing on the sky portion of FMB image 01431, it is evident that the semantic segmentation result obtained from our fusion method is closer to the visible image. Similarly, on the vehicle portion of FMB image 01382, our fused image achieves a more complete segmentation result.

\noindent \textbf{Detection Performance.}
We conducted object detection experiments on the MSRS detection dataset using the pre-trained YOLOv8 model~\citep{jiang2022review,hussain2024yolov1}. Before testing, we used GroundedSAM to segment the detection set to obtain prompt masks. We provide several visual examples in Fig.~\ref{fig:MDdetect} to demonstrate the effectiveness of our fusion algorithm in enhancing object detection performance. In the MSRS scene 583, the detector fails to detect a car from the Visible, LDFusion, CDDFuse, PSFusion, and SDCFusion images. While SwinFuse and DIDFuse produce error detections of an additional person, SeAFusion incorrectly identifies an extra vehicle. In DroneVehicle image 151, false detections are observed in the infrared, LDFusion, NestFuse, and DIDFuse images. Additionally, missing detections occur in the visible images, as well as with SwinFuse, NestFuse, DIDFuse, and PSFusion. In these two sets of examples, only our method successfully avoids both error detections and missing detections. This demonstrates that our method provides sufficient semantic information for high-level vision tasks.

\noindent \textbf{Comparison of CtrlFuse with and without Mask Prompts.}
Our method is an interactive and controllable fusion algorithm. During the inference phase, we employ GroundedSAM to interactively generate mask prompts to guide the model learning as shown in Fig.~\ref{fig:Ctrltest}, while also supporting prompt-free input. To verify the contribution of our method with prompt masks in downstream tasks, we compare its performance with and without prompt masks on semantic segmentation and object detection.

\begin{figure*}[t]
	\centering
	\includegraphics[width=1.0\textwidth]{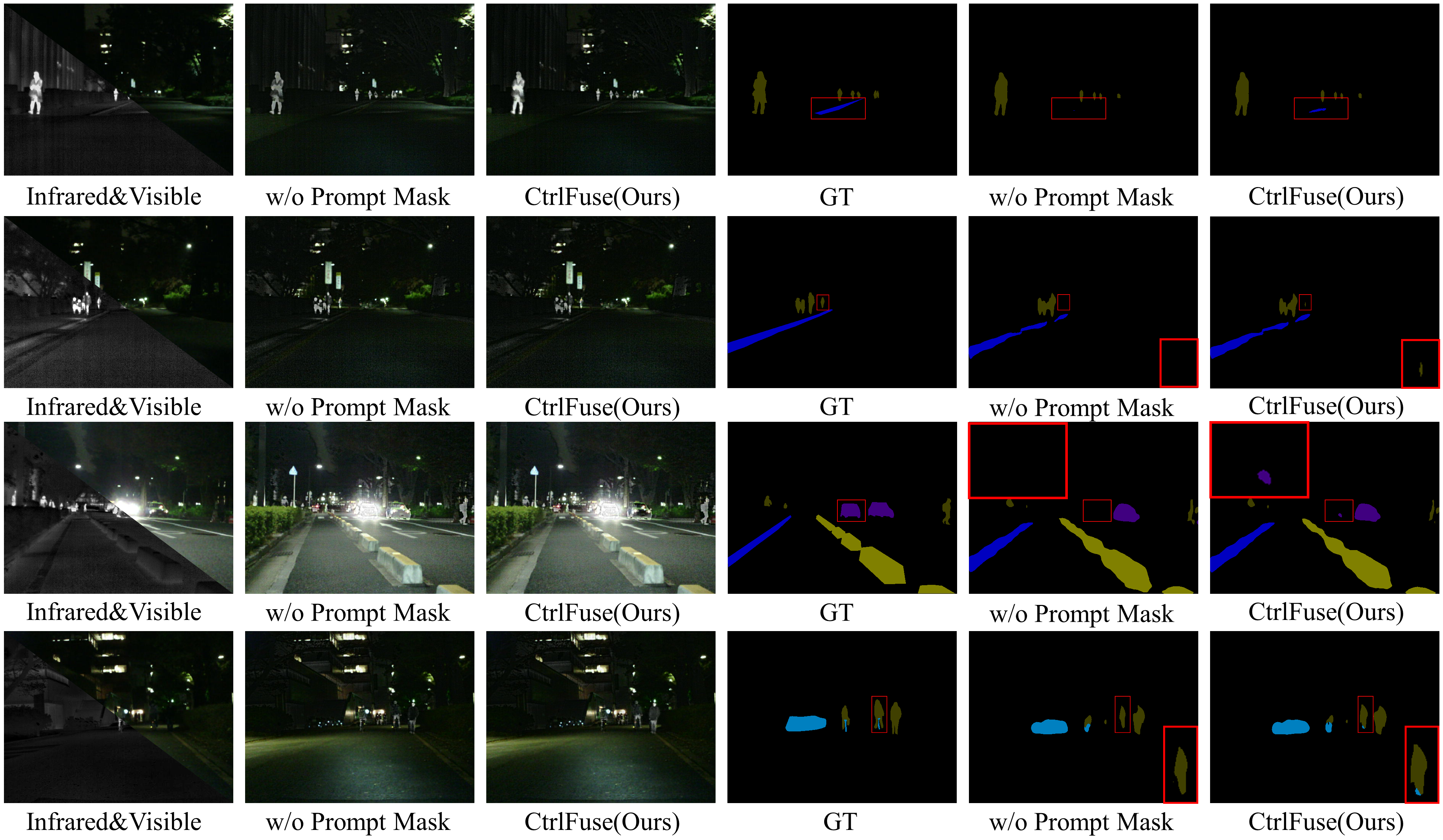} 
	\caption{Qualitative comparisons of segmentation performance with and without mask prompts on the MSRS dataset.}
	\label{fig:mask_prompt_segcomparison}
\end{figure*}

In Table~\ref{tab:mask_prompt_segcomparison}, we present the semantic segmentation test metrics of the method without Prompt Mask and our method on the MSRS dataset. Following the same semantic segmentation experimental setup as before, we conduct testing using retrained DeepLabV3+ semantic segmentation models. Our method achieves improvements over the no-prompt approach in six categories and demonstrates superiority in mIoU. To provide a more intuitive comparison, we also offer visual comparisons as shown in Fig~\ref{fig:mask_prompt_segcomparison}. We provide a visual comparison of segmentation results in four scenarios. In the first row (MSRS 00862N), our method successfully segments the ``curve" region, while the no-prompt approach fails. In the second row (MSRS 00874N), the distant ``person" is not detected by the no-prompt method, whereas our method manages to segment a partial region of the person. In the third row (MSRS 00931N), the no-prompt method loses the semantic information of the ``car" due to overexposure, while our method obtains an incomplete but recognizable segmentation mask. In the fourth row, the target is a person riding a bicycle; the no-prompt method fails to segment the bicycle, while our method successfully captures it. These visual comparisons further demonstrate that our prompt-guided fusion method is more effective in preserving semantic information.

\begin{table}[t]
\centering
\setlength{\tabcolsep}{1mm}
\begin{tabular}{l c c c c c}
\toprule
& \multicolumn{5}{c}{\textbf{mAP on DroneVehicle Dataset}} \\
\cmidrule(lr){2-6}
\multirow{-2}{*}{Metrics} & car & truck & bus & freight car & All \\
\midrule
\textbf{AP@0.5} & & & & & \\
\quad w/o Mask Prompt  & 0.929 & 0.591 & 0.674 & 0.413 & 0.652 \\
\quad CtrlFuse(Ours)    & \textbf{0.952} & \textbf{0.639} & \textbf{0.708} & \textbf{0.533} & \textbf{0.708} \\
\addlinespace[3pt]
\textbf{AP@0.7} & & & & & \\
\quad w/o Mask Prompt  & 0.826 & 0.560 & 0.642 & 0.405 & 0.608 \\
\quad CtrlFuse(Ours)    & \textbf{0.877} & \textbf{0.597} & \textbf{0.708} & \textbf{0.519} & \textbf{0.675} \\
\addlinespace[3pt]
\textbf{AP@[0.5:0.95]} & & & & & \\
\quad w/o Mask Prompt  & 0.618 & 0.420 & 0.454 & 0.315 & 0.451 \\
\quad CtrlFuse(Ours)    & \textbf{0.651} & \textbf{0.521} & \textbf{0.521} & \textbf{0.409} & \textbf{0.525} \\
\bottomrule
\end{tabular}
\caption{Object detection performance comparison on the DroneVehicle dataset with and without mask prompts. Metrics include mAP@0.5, mAP@0.7, and mAP@[0.5:0.95] for car, truck, bus, freight car, and overall (All) classes. \textbf{Bold} indicates the best.}
\label{tab:mask_prompt_detectcomparison}
\end{table}

\begin{figure}[t]
	\centering
	\includegraphics[width=0.5\textwidth]{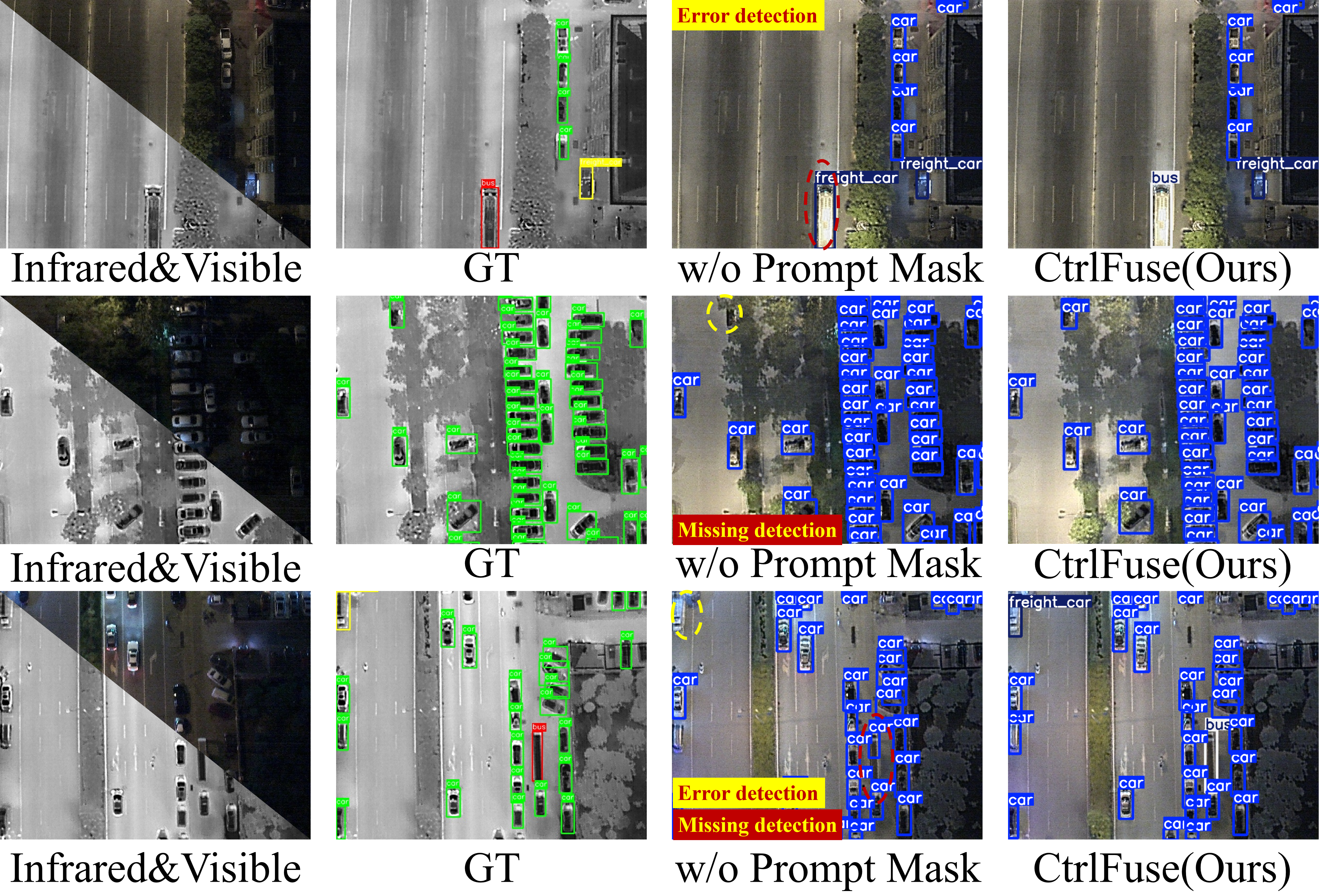} 
	\caption{Qualitative comparisons of object detection performance with and without mask prompts on the DroneVehicle dataset.}
	\label{fig:mask_prompt_detectcomparison}
\end{figure}

In Table~\ref{tab:mask_prompt_detectcomparison}, we present the object detection test metrics of the method without Prompt Mask and our method on the DroneVehicle dataset. Our method performs better than the method without Mask Prompt across all metrics. In addition, we have also provided qualitative comparisons as shown in Fig.~\ref{fig:mask_prompt_detectcomparison}.In the first row, scene 423 of DroneVehicle, the method without Prompt Mask misclassifies a ``bus" as a ``freight car". In the second row, scene 1064 of DroneVehicle, the prompt-free method fails to detect a car. In the third row, scene 1373, the prompt-free method misses a freight car and misclassifies a bus as a car. Through quantitative and qualitative analysis, the fusion method guided by the mask prompt significantly outperforms the non-prompted approach in object detection. This further demonstrates that prompt guidance enables the algorithm to focus more on our designated regions of interest and enriches the detailed information.

\begin{table*}[h]
\centering
\small
\begin{tabular}{l *{9}{c} c}
\toprule
Method & Background & Car & Person & Bike & Curve & \begin{tabular}{@{}c@{}}Car\\Stop\end{tabular} & Guardrail & \begin{tabular}{@{}c@{}}Color\\Tone\end{tabular} & Bump & mIoU \\
\midrule
SAGE & 0.970 & 0.777 & 0.498 & 0.743 & 0.613 & 0.797 & 0.985 & \textbf{0.812} & \textbf{0.945} & 0.793 \\
CtrlFuse (Ours) & \textbf{0.971} & \textbf{0.781} & \textbf{0.513} & \textbf{0.744} & \textbf{0.618} & \textbf{0.802} & 0.985 & 0.808 & 0.944 & \textbf{0.796} \\
\bottomrule
\end{tabular}
\caption{Quantitative comparison on semantic segmentation task with SAGE method.\textbf{Bold} indicates the best.}
\label{tab:SAGEseg}
\end{table*}

\subsection{Comparison with SAM-based Fusion Method SAGE}
Wu et al.~\cite{wu2025every} proposed a fusion method that utilizes semantic priors from the SAM for IVIF. In this subsection, we conduct both qualitative and quantitative analyses in comparison with SAGE.

\begin{figure}[!h]
	\centering
	\includegraphics[width=0.5\textwidth]{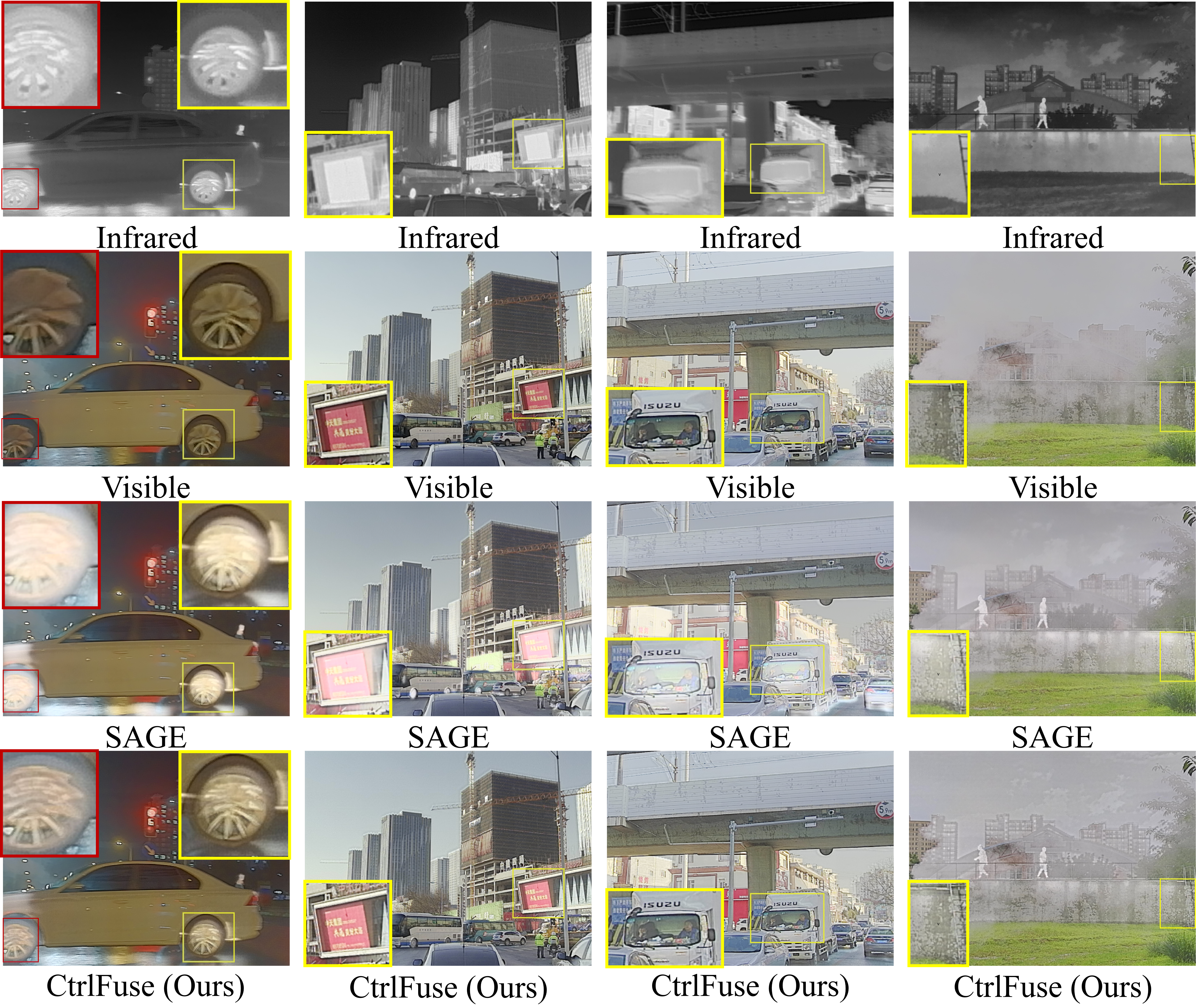} 
	\caption{Qualitative comparisons with SAGE on representative images selected from the FMB dataset.}
       \label{fig:SAGEFMB}
\end{figure}

\begin{figure}[!h]
	\centering
	\includegraphics[width=0.5\textwidth]{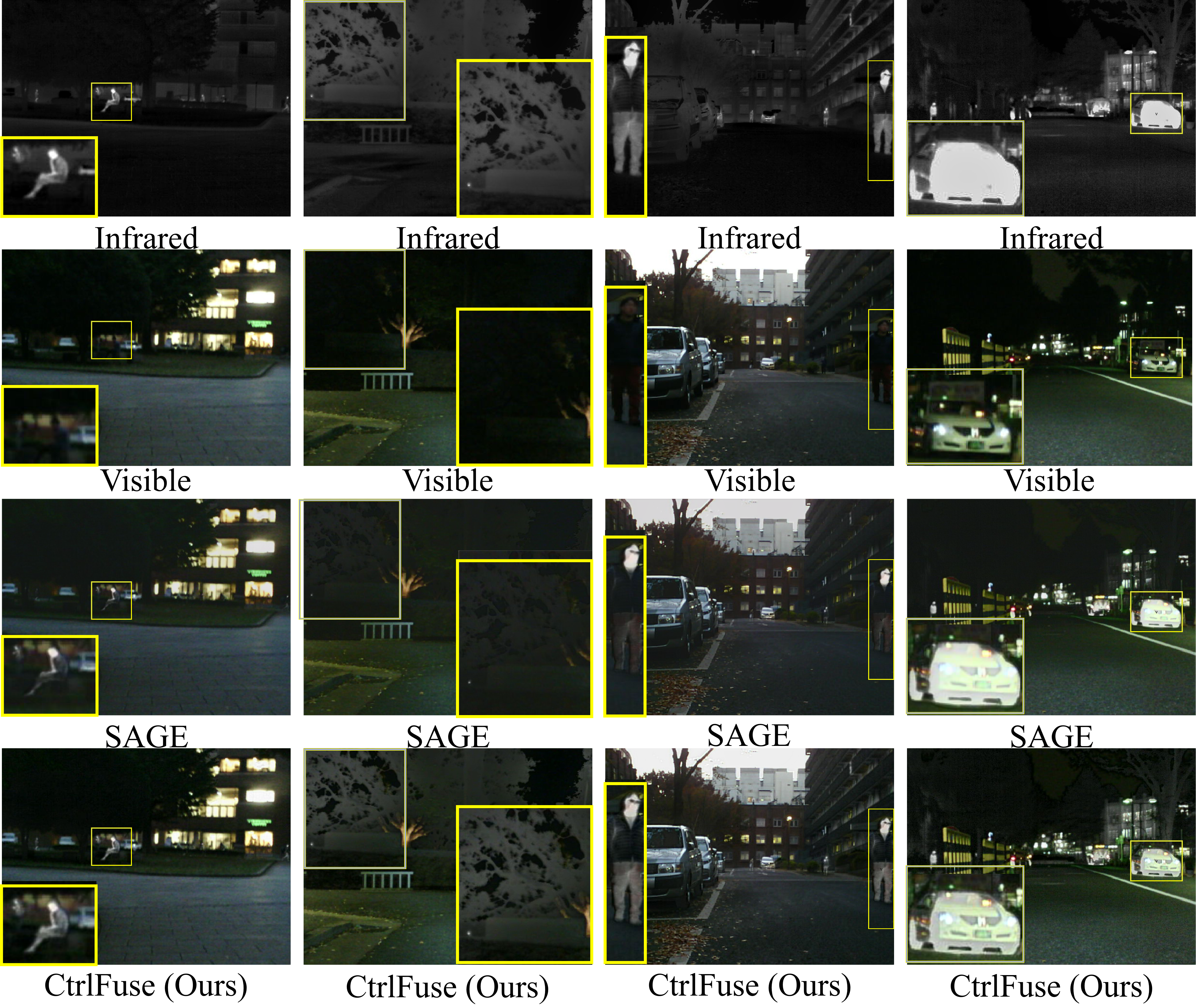} 
	\caption{Qualitative comparisons with SAGE on representative images selected from the MSRS dataset.}
       \label{fig:SAGEMSRS}
\end{figure}

\begin{table}[h]
\centering
\setlength{\tabcolsep}{2pt}
\begin{tabular}{l r r r r r r}
\toprule
Method & MSE & PSNR & $Q_{\text{abf}}$ & $N_{\text{abf}}$ & SSIM &SCD \\
\midrule
SAGE & 0.09 & 58.844 & 0.336 & 0.035 & 0.677 & 1.413 \\
CtrlFuse(Ours) & \textbf{0.063} & \textbf{60.317} & \textbf{0.496} & 0.035 & \textbf{0.779} & \textbf{1.552} \\
\bottomrule
\end{tabular}
\caption{Comparison of SAGE and CtrlFuse Methods on the DroneVehicle dataset.Bold indicates the optimal metric values.}
\label{tab:SAGEdrone}
\end{table}

\begin{figure}[!h]
	\centering
	\includegraphics[width=0.5\textwidth]{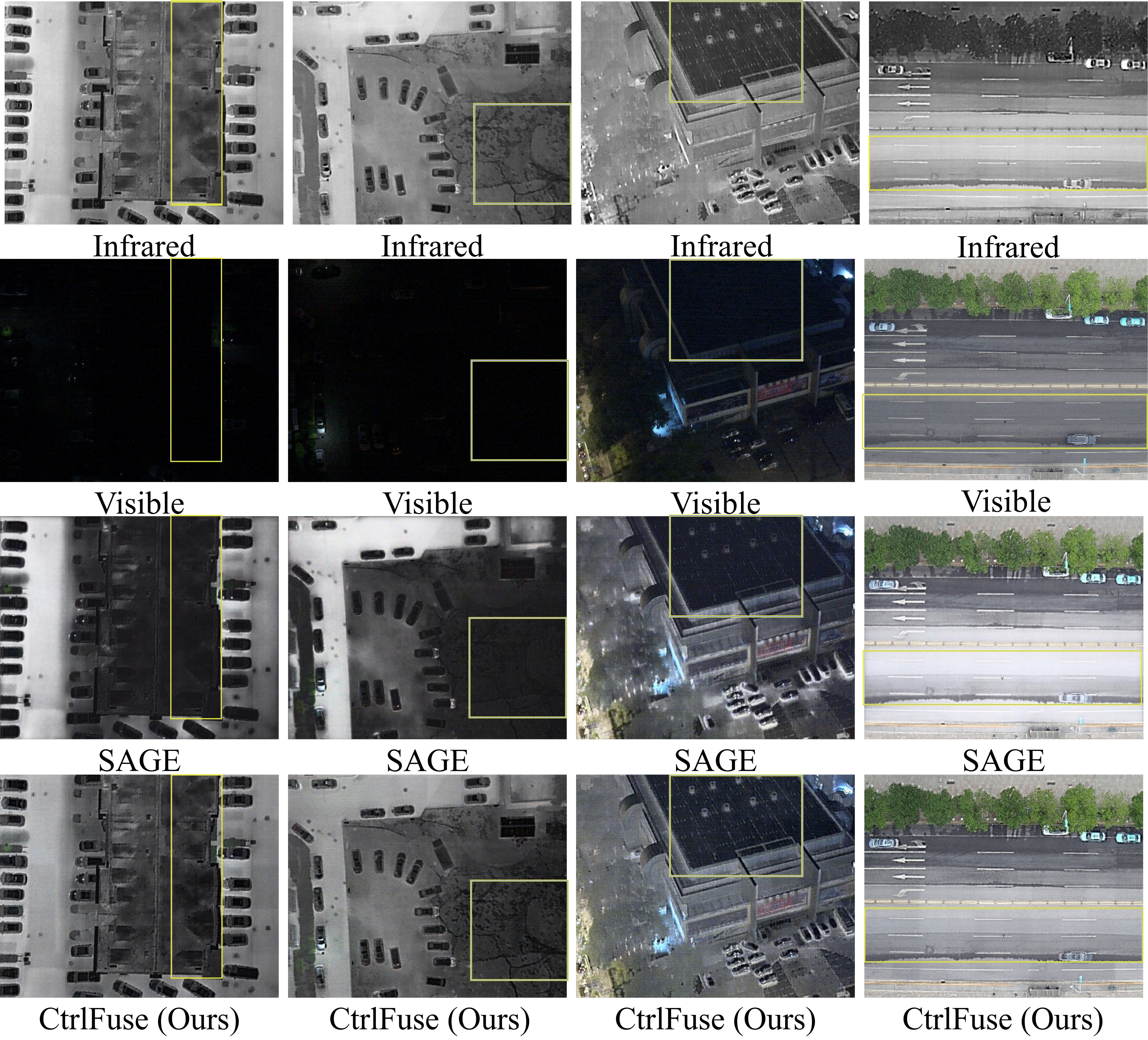} 
	\caption{Qualitative comparisons with SAGE on representative images selected from the DroneVehicle dataset.}
       \label{fig:SAGEdrone}
\end{figure}

\noindent \textbf{Comparison on the FMB Dataset.}
Qualitative comparisons were performed on the FMB dataset, as illustrated in Fig.~\ref{fig:SAGEFMB}. In the first column of images, we use red and yellow rectangles to mark the two foreground objects, respectively, and magnify these areas for easier comparison. We focus on the details of car hubs. Infrared images provide rich brightness and texture, whereas visible images mainly capture texture. In SAGE's fused results, texture is obscured by dominant brightness information, causing blurry edges and reduced detail. Our method better balances the integration of this information, preserving texture clarity and edge sharpness. In the second to fourth columns, we use yellow rectangular boxes to highlight the targets and zoom in on these areas for detailed comparison. These columns respectively focus on a signboard (second), a person within a truck (third), and wall stains (fourth). The cases presented in the second to fourth columns represent daytime scenarios, in which infrared features play a more supplementary role. Nevertheless, the SAGE fusion method introduces an overemphasis on infrared characteristics, resulting in significant texture degradation. Our approach, by comparison, achieves a more balanced integration that effectively retains the textual information.

\noindent \textbf{Comparison on the MSRS Dataset.}
We conducted qualitative analyses on the MSRS dataset, as shown in Fig.~\ref{fig:SAGEMSRS}. The first and third columns emphasize the human figures within the scenes. In such cases, the thermal information corresponding to ``person" becomes a key focus of analysis. As illustrated in the figure, the SAGE fusion method introduces extraneous information from the visible images into the 'person' regions in both examples. This interference leads to a degradation of thermal detail, making the thermal information less prominent. In contrast, our method effectively preserves the thermal characteristics. The second and fourth columns highlight the textural details of branches and cars, respectively. In the second column, SAGE places excessive emphasis on the visible image, causing the branch textures to become indistinct. In the fourth column, however, an overemphasis on the infrared image leads to the attenuation of visible information, which results in a loss of clarity in the car’s front section.Relative to SAGE, our approach excels in balancing the image features from both infrared and visible modalities. By appropriately combining the thermal information from infrared images with the detailed textures from visible images, our method retains clarity and detail in both aspects.

\begin{table}[t]
\centering
\setlength{\tabcolsep}{1mm}
\begin{tabular}{l c c c c c}
\toprule
& \multicolumn{5}{c}{\textbf{Object Detection Performance}} \\
\cmidrule(lr){2-6}
\multirow{-2}{*}{Methods} & car & truck & bus & freight car & All \\
\midrule
\textbf{AP@0.5} & & & & & \\
\quad SAGE  & 0.922 & \textbf{0.659} & 0.591 & 0.430 & 0.651 \\
\quad CtrlFuse (Ours) & \textbf{0.952} & 0.639 & \textbf{0.708} & \textbf{0.533} & \textbf{0.708} \\
\addlinespace[3pt]
\textbf{AP@[0.5:0.95]} & & & & & \\
\quad SAGE  & 0.627 & 0.491 & 0.419 & 0.337 & 0.468 \\
\quad CtrlFuse (Ours) & \textbf{0.651} & \textbf{0.521} & \textbf{0.521} & \textbf{0.409} & \textbf{0.525} \\
\bottomrule
\end{tabular}
\caption{Object detection performance comparison with SAGE method. \textbf{Bold} indicates superior performance.}
\label{tab:SAGEdetect}
\end{table}

\begin{table}[h]
\centering
\setlength{\tabcolsep}{2.5pt}
\footnotesize
\begin{tabular}{lcccccc}
\toprule
Method & \rotatebox{0}{MSE} & \rotatebox{0}{PSNR } & \rotatebox{0}{Qabf } & \rotatebox{0}{Nabf } & \rotatebox{0}{SSIM} & \rotatebox{0}{SCD } \\
\midrule
Reference Fusion & 0.046 & 62.96 & 0.717 & 0.029 & 0.917 & 1.46 \\
CtrlFuse (Ours) & \textbf{0.043} & \textbf{63.292} & \textbf{0.719} & \textbf{0.024} & \textbf{0.925} & \textbf{1.522} \\
\bottomrule
\end{tabular}
\caption{Quantitative comparison between the reference fusion and our final output.}
\label{tab:ref}
\end{table}

\noindent \textbf{Comparison on the DroneVehicle Dataset.}
The quantitative results on the DroneVehicle dataset are shown in Table~\ref{tab:SAGEdrone}. Our method outperforms SAGE on five evaluation metrics, while achieving comparable performance on the sixth. The high PSNR and SSIM values of fused images indicate superior visual quality and detail preservation, while lower MSE ensures accurate information retention. Enhanced $Q_{\text{abf}}$ scores improve the clarity of blurred frames, aiding reliable tracking and detection. Improved SCD performance allows precise identification of spectral changes, beneficial for distinguishing subtle environmental differences.

To clearly distinguish the object regions, we annotate them with a yellow rectangular box in Fig.~\ref{fig:SAGEdrone}. In the first three columns of the comparison images, SAGE's inability to balance primary and auxiliary features leads to amplified interference. This results in the fused images being biased towards visible image information, causing important features to be lost. Conversely, as shown in the final set of comparison images, SAGE tends to over-enhance the brightness information derived from the infrared input, which results in the suppression of road landmarks and surface textures due to excessive brightness. In contrast, our method avoids over-amplifying information from a single modality, which can lead to information loss. Instead, it effectively balances information from both modalities, ensuring that no critical details are obscured.

\begin{figure}[t]
	\centering
	\includegraphics[width=0.5\textwidth]{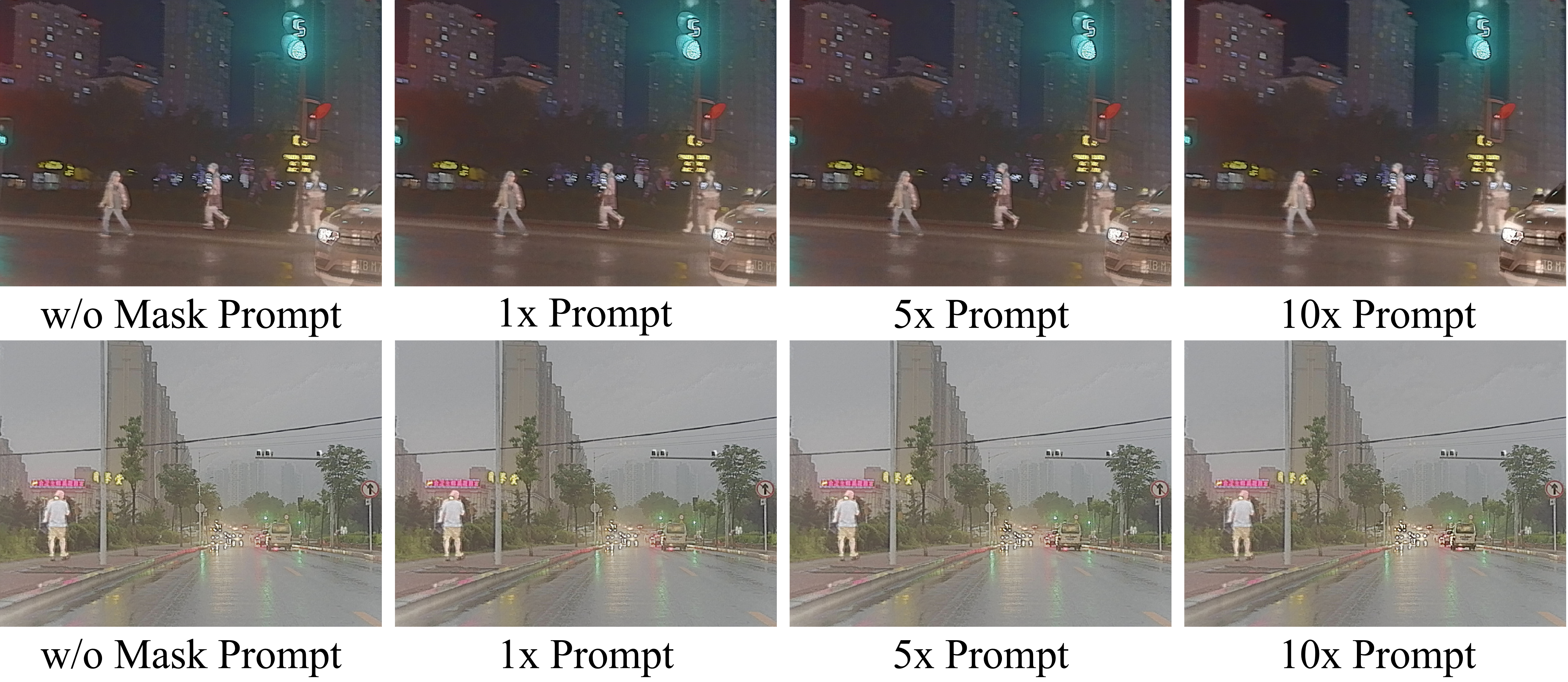} 
	\caption{The fusion results obtained with the \textbf{``car" mask prompt} at varying control parameter levels are shown from left to right: no mask prompt, 1× prompt, 5× prompt, and 10× prompt.}
       \label{fig:carprompt}
\end{figure}

\begin{figure}[t]
	\centering
	\includegraphics[width=0.5\textwidth]{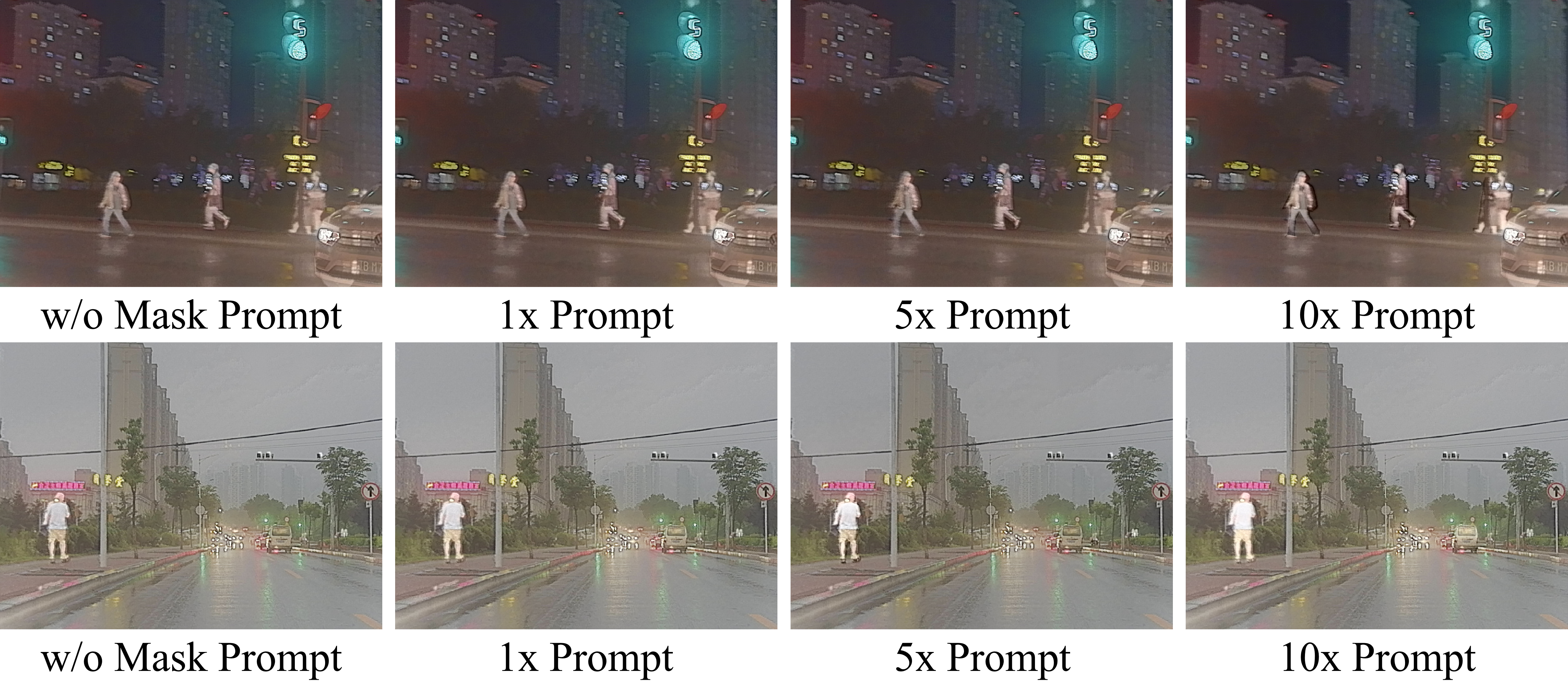} 
	\caption{The fusion results obtained with the \textbf{``person" mask prompt} at varying control parameter levels are shown from left to right: no mask prompt, 1× prompt, 5× prompt, and 10× prompt.}
       \label{fig:personprompt}
\end{figure}

\begin{figure}[t]
	\centering
	\includegraphics[width=0.5\textwidth]{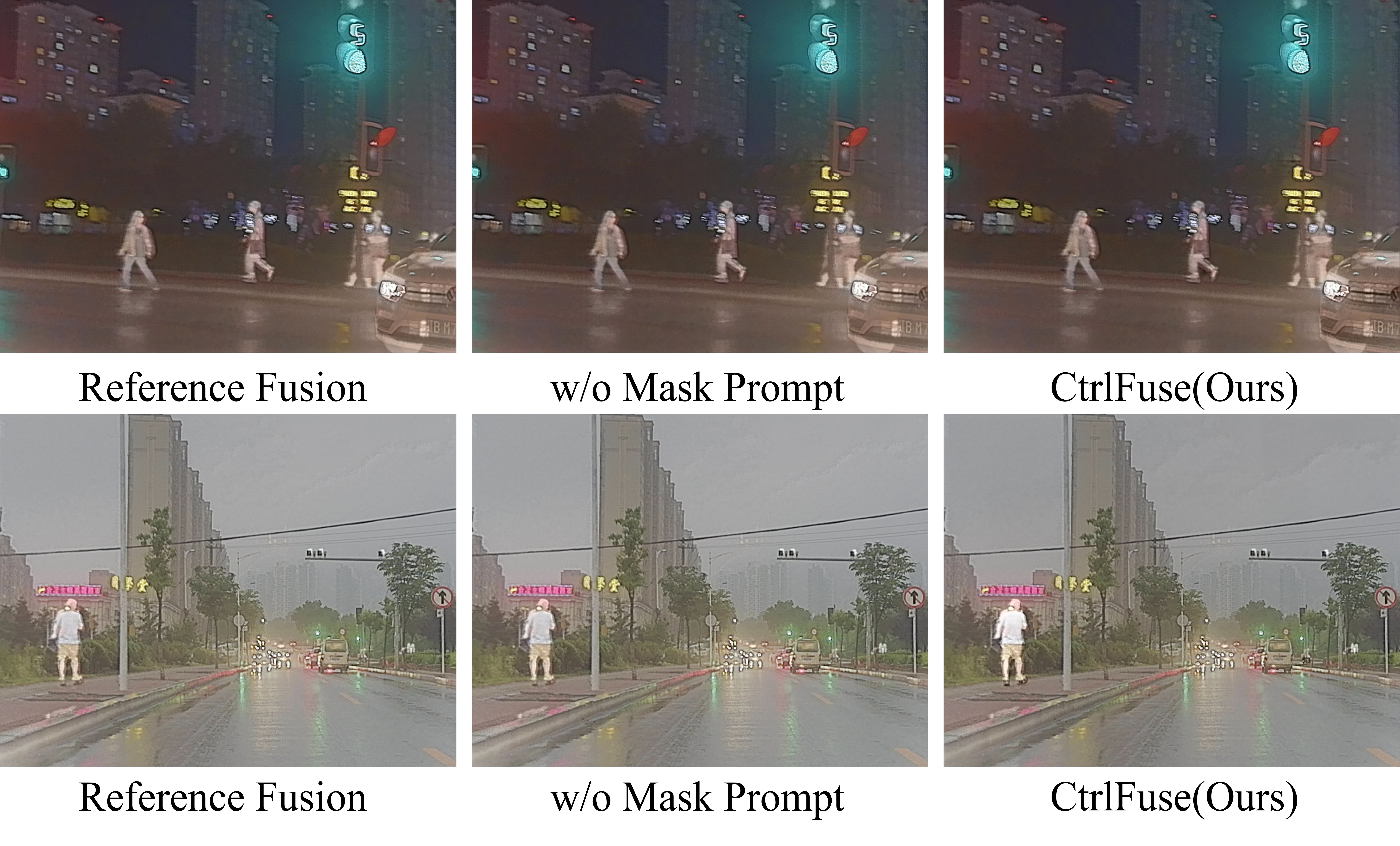} 
	\caption{Qualitative comparison of the reference fusion image, prompt-free result, and our prompt-based fusion result.}
       \label{fig:ref}
\end{figure}

\noindent \textbf{Comparisons on Semantic Segmentation.}
To further validate the semantic extraction capability of our method, we conducted comparative experiments with SAGE, another SAM-based approach, on the semantic segmentation task. Quantitative results are presented in Table~\ref{tab:SAGEseg}.The superior performance on the majority of classes and in mIoU demonstrates an enhanced utilization of SAM's semantic comprehension capabilities by our method.

\noindent \textbf{Comparisons on Object Detection.}
To further validate the effectiveness of our method on downstream tasks, we compare its object detection performance with SAGE. As quantitatively demonstrated in Table~\ref{tab:SAGEdetect}, our approach achieves superior results across nearly all evaluation metrics. This outcome further confirms the enhanced capability of our method in leveraging semantic information for improved object detection.

\subsection{Controllable Fusion Details}
During testing, we can alter the fused image by controlling the fusion parameters, allowing the highlighted regions to appear at different levels of prominence. In the Fig.~\ref{fig:carprompt} and Fig.~\ref{fig:personprompt}, we demonstrate examples of the ``car" and ``person" classes using ``no prompt", ``1x prompt", ``5x prompt", and ``10x prompt" to illustrate the effects of prompt features. As can be seen in the two examples, the prompted regions in the fused images become increasingly prominent from left to right. In the case of the ``car" example as Fig.~\ref{fig:carprompt}, increasing the intensity of the prompt feature primarily enhances the visible characteristics, resulting in clearer texture details. In the first row, as the intensity of the prompt is increased, the vehicle in the frame contains richer texture detail (sharper license plate numbers), and in the second row, the detail of the vehicle in the far distance is sharper. In the ``person" example illustrated in Fig.~\ref{fig:personprompt}, the fusion results progressively favor infrared information as we move from left to right. As the intensity of the prompt increases, the pedestrians in the first and second rows of images possess more significant contrast in nighttime conditions. In the two experiments mentioned above, we further validated that the fusion algorithm we proposed features interactive controllability.

\subsection{Comparison of Fusion Result with the Reference Fusion Image}
The intermediate fusion result $I_{\text{ref}}$ is generated during the process and serves as a reference fusion image for subsequent modules. Its primary role is to provide a query mechanism that guides the entire fusion process toward the desired optimization direction. We conducted both quantitative and qualitative comparisons between $I_{\text{ref}}$ and $I_{\text{F}}$, with the results detailed in Table~\ref{tab:ref} and Fig.~\ref{fig:ref}, respectively. As can be observed from the qualitative comparison figures, the reference fusion map exhibits similarity to the fusion results without mask prompt.

\end{document}